\documentclass[titlepage,11pt]{article}
\usepackage[dvips]{graphicx}
\usepackage[myheadings]{fullpage}
\usepackage{pmetrika}

\usepackage{submit}

\usepackage{amsmath, amsfonts}
\usepackage{bm} % bold
\usepackage{float} % table head
\usepackage{color} % color
\usepackage{multirow}
\usepackage{booktabs}
\usepackage{apacite}
\usepackage{natbib} % citation formats

\usepackage[ruled, vlined]{algorithm2e}
\SetKwInput{KwData}{Input}
\SetKwInput{KwResult}{Output}

\DeclareMathOperator*{\argmax}{arg\,max}

\newcommand{\blue}[1]{{\color{black}#1}}

\graphicspath{{./figs/}}
\DeclareGraphicsExtensions{.pdf,.eps}
\begin{document}

%% ITEM 6 [See the "howto.tex" file.]
\begin{titlepage}

\title{}

%%Disable \markright for your submission,
%%if it includes any author names.
%%Note: The "submit" package includes a running header.
%\markright{\MakeLowercase{\textsc{}}}

\author{Xiao Li}
\affil{Department of Educational Psychology\break University of Illinois at Urbana-Champaign}

\author{Hanchen Xu}
\affil{Department of Electrical and Computer Engineering\break University of Illinois at Urbana-Champaign}

\author{Jinming Zhang}
\affil{Department of Educational Psychology\break University of Illinois at Urbana-Champaign}

\author{Hua-hua Chang}
\affil{Department of Educational Studies\break Purdue University}

\vspace{\fill}\centerline{\today}\vspace{\fill}

%\comment{This research was funded by .}
%\thanks{I would like to thank .}
%\linespacing{1}
%\contact{Correspondence should be sent to\\
%
%\noindent E-Mail: xiaoli20@illinois.edu \break
%\noindent Phone: \break
%\noindent Fax: \break
%\noindent Website:  }

\end{titlepage}

%% ITEM 7 [See the "howto.tex" file.]
\setcounter{page}{2}
\vspace*{2\baselineskip}

\RepeatTitle{Deep Reinforcement Learning for Adaptive Learning Systems}\vskip3pt

\linespacing{1.5}
%% ITEM 8 [See the "howto.tex" file.]
\abstracthead
\begin{abstract}
In this paper, we formulate the adaptive learning problem---the problem of how to find an individualized learning plan (called policy) that chooses the most appropriate learning materials based on learner's latent traits---faced in adaptive learning systems as a Markov decision process (MDP).
We assume latent traits to be continuous with an unknown transition model.
We apply a model-free deep reinforcement learning algorithm---the deep Q-learning algorithm---that can effectively find the optimal learning policy from data on learners' learning process without knowing the actual transition model of the learners' continuous latent traits.
To efficiently utilize available data, we also develop a transition model estimator that emulates the learner's learning process using neural networks.
The transition model estimator can be used in the deep Q-learning algorithm so that it can more efficiently discover the optimal learning policy for a learner.
Numerical simulation studies verify that the proposed algorithm is very efficient in finding a good learning policy, especially with the aid of a transition model estimator, it can find the optimal learning policy after training using a small number of learners.

\begin{keywords}
adaptive learning system, transition model estimator, Markov decision process, deep reinforcement learning, deep Q-learning, neural networks, model-free
\end{keywords}
\end{abstract}\vspace{\fill}\pagebreak

%% ITEM 8 [See the "howto.tex" file.]

%%%%%%%%%%%%%%%%%%%%%%%%%%%%%%%%%%%%%%%%%%%%%%%%%%%%%%%%%%%%%%%%%%%%%%
\section{Introduction}

In a traditional classroom, a teacher uses the same learning material (e.g. textbook, instruction pace, etc.) for all students. 
However, the selected material may be too hard for some students and too easy for some other students. 
Further, some students may take longer time in learning than the others. 
Such a learning process may not be efficient. 
These issues can be solved if the teacher can make an individualized learning plan for each individual student: Select an appropriate learning material according to each student' ability and let a student learn at her/his own pace. 
Considering that a very low teacher-student ratio is required, such an individualized adaptive learning plan may be too expensive to be applied to all students. As such, adaptive learning systems are developed to provide individualized adaptive learning for all students/learners. 
In particular, with the fast growth of digital platforms, globally integrated resources, and machine learning algorithms, the adaptive learning systems are becoming increasingly more affordable, applicable, and efficient \citep{zhang2016smart}.

An adaptive learning system---also referred to as a personalized/individualized learning or intelligent tutoring system---aims at providing a learner with optimal and individualized learning experience or instructional materials so that the learner can reach a certain achievement level in a shortest time or reach as high as possible an achievement level in a fixed period of time.  
First, learners' historical data are used to estimate her/his proficiency. 
Then, according to the level of her/his proficiency, the system selects the most appropriate learning material for the learner. 
After the learner finishes the learning material, an assessment is given to the learner and her/his proficiency level is updated and is used by the adaptive learning system to choose the next most appropriate learning material for the learner. 
Such process repeats until the learner achieves a certain proficiency level. 

In previous studies, the proficiencies or latent traits were typically characterized as vectors of binary latent variables \citep{chen2018recommendation,li2018optimal,tang2019reinforcement}. 
However, it is important to consider the granularity of the latent traits in a complicated learning and assessment environment in which a knowledge domain consists of several fine-grained abilities. 
In some cases, it would be too simple to model learners' abilities as mastery or non-mastery. 
For example, when an item is designed to measure several latent traits and a learner regarded as mastering all related traits of the item cannot be assured to answer the item correctly. 
A possible reason is that the so-called mastery is not full mastery of a latent trait.
By measuring learners' traits as continuous scales, the adaptive learning system can be designed to help learners to learn and improve until they reach the target levels of certain abilities so that the learners can achieve target scores in assessments.
Especially in practice, most assessments are designed to measure learners' latent traits \citep{mcglohen2008combining}.
In such scenarios, it is better to use a continuous scale to measure the latent traits as the item response theory (IRT) does.
In this paper, we will develop an adaptive learning system that estimate learners' abilities using measurement models in order to provide them with most appropriate materials for further improvements.

Existing research studies have focused on modeling learners' learning paths \citep{chen2018hidden,wang2018tracking}, accelerating learners' memory speed \citep{reddy2017accelerating}, providing model-based sequence recommendation \citep{chen2018recommendation,lan2016contextual,xu2016personalized}, tracing learners' concept knowledge state transitions over time \citep{lan2014time}, and \blue{selecting materials for learners optimally based on model-free algorithms \citep{li2018optimal,tang2019reinforcement}.}
However,  \blue{explicit models are typically needed to characterize learners' learning progresses in these studies.
While there exist research studies that aim to find the optimal learning strategy/plan (called \textit{policy} in the rest of the paper) which chooses the most appropriate learning materials for learners using model-free algorithms, they all assume discrete latent traits.}
In addition, when the number of learners is too small for the system to learn an optimal policy, these algorithms are not applicable.
This paper studies the important, yet less addressed adaptive learning problem---the problem of finding the optimal learning policy---based on continuous latent traits, and applies machine learning algorithms to deal with the tackle challenges such as only a small number of learners available in historical data.

In this paper, we formulate the adaptive learning problem as a Markov decision process (MDP), in which the state is the (continuous) latent traits of a learner, the action is the (discrete) learning material given to the learner.
Yet, the state transition model is unknown in practice, thus making the MDP unsolvable using conventional model-based algorithms such as the value iteration algorithm \citep{sutton2018reinforcement}. 
To solve the issue, we apply a model-free deep reinforcement learning (DRL) algorithm, the so-called deep Q-learning algorithm, to search for the optimal learning policy.
The model-free DRL algorithm is a class of machine learning algorithms that solve an MDP by learning an optimal policy represented by neural networks from a sequence of state transitions directly when the transition model itself is are unknown \citep{franccois2018introduction}.
DRL algorithms have been widely applied in solving a variety of problems in many different fields such as playing Atari games \citep{mnih2015human}, bidding and pricing in electricity market \citep{xu2019deep}, manipulating robotics \citep{gu2017deep}, and localizing objects \citep{caicedo2015active}.
We refer interested readers to \citet{franccois2018introduction} for a detailed review on the theories and applications of DRL.
Therefore, the adaptive learning system is embedded with the well-developed measurement models and the model-free DRL algorithm so as to be more flexible.

However, a deep Q-learning algorithm typically requires a large amount of state transition data so as to find an optimal policy, which may be difficult to obtain in practice. 
To cope with the challenge of insufficient state transition data, we develop a transition model estimator that emulates the learner's learning process using neural networks. 
The transition model that is fitted using available historical transition data can be used in the deep Q-learning algorithm to further improve its performance with no additional cost. 

The purpose of this paper is to develop a \textit{fully adaptive} learning system in which (i) the learning material given to a learner is based on her/his continuous latent traits that indicate the levels of certain abilities, and (ii) the learning policy that maps the learner's latent traits to the learning materials is found adaptively with minimal assumption on the learners' learning process.
First, an MDP formulation for the adaptive learning problem by representing latent traits in a continuum is developed.
Second, a model-free DRL algorithm---the deep Q-learning algorithm---is applied, to the best of our knowledge, for the first time, in solving the adaptive learning problem.
Third, a neural network based transition model estimator is developed, which can greatly improve the performance of the deep Q-learning algorithm when the number of learners is inadequate.
Last, some interesting simulation studies are conducted to serve as demonstration cases for the development of adaptive learning systems.
\color{black}

The remainder of this paper is organized as follows.
In the Preliminaries section, we briefly review measurement models and make some assumptions on the adaptive learning problem.
In the Adaptive Learning Problem section, we introduce the conventional adaptive learning systems and develop a MDP formulation for the adaptive learning problem.
Then, we apply the deep Q-learning algorithm to solve the MDP in the Optimal Learning Policy Discovery Algorithm section, where a transition model estimator that emulates the learners is also developed. 
Two simulation studies are conducted in the Numerical Simulation section and some concluding remarks are made at the end of the paper.

% define adaptive learning system and the adaptive learning problem---the problem of finding a learning policy that chooses learning materials based on a learner's latent traits---faced in adaptive learning systems

\section{Preliminaries}

In this section, we give a brief introduction on measurement models for continuous latent traits, which is an important component in adaptive learning systems.
The representation of learners' latent traits and assumptions on them are also presented.

\subsection{Measurement Models}

In an adaptive learning system, a test is given to a learner/student after each learning cycle. 
The learner's responses to the test items are collected by the system and her/his latent traits are estimated using measurement models, specifically IRT models \citep{rash1960probabilistic,lord1968statistical}. 

An appropriate IRT model needs to be chosen based on the test's features such as the test's dimensional structure \citep{zhang2013procedure} and its response categories. 
To be more specific, in the case when item responses are recorded as binary values indicating correct or incorrect answers, the test that evaluates only one latent trait will use the unidimensional item response theory IRT models \citep{rash1960probabilistic, birnbaum1968some, lord1980application}, whereas tests that associate more than one trait will use the multidimensional item response theory (MIRT) models \citep{reckase1972development, mulaik1972mathematical, sympson1978model, whitely1980multicomponent}. 
When item responses have more than two categories, polytomous IRT models such as the partial credit model \citep{masters1982rasch}, the generalized partial credit model \citep{muraki1992generalized}, and the graded response model \citep{samejima1969estimation} are used for unidimensional case. 
Their extensions can be applied in multidimensional cases. 

The basic representation of an IRT model is expressed as
\begin{align} \label{eq1}
	\mathbb{P}(U = u | \bm{\theta}) = f(\bm{\theta}, \bm{\eta}, u),
\end{align}
where $\mathbb{P}$ denotes probability, $U$ is a random variable representing the score on the test item, $u$ is the possible value of $U$, $\bm{\theta}$ is a vector of parameters describing the learner's latent traits, $\bm{\eta}$ is a vector of parameters indicating the characteristic of the item, and $f$ denotes a function that maps $\bm{\theta}, \bm{\eta}, u$ to a probability in $[0, 1]$.
As pointed out in \citet{ackerman2003using}, many educational tests are inherently multidimensional. 
Therefore, we will use the MIRT as the intrinsic model to build up the adaptive learning system.
As an illustration, the multidimensional two-parameter logistic IRT (M2PL) model is given by
\begin{align} \label{eq2}
	\mathbb{P}(U_{ij}=1|\bm{\theta}_i, \bm{a}_j, d_j)=\frac{e^{\bm{a}_j^\top \bm{\theta}_i + d_j}}{1+e^{\bm{a}_j^\top \bm{\theta}_i+d_j}},
\end{align}
where $U_{ij}$ is the response given by $i^{th}$ test taker to $j^{th}$ item, 
$\bm{\theta}_i= [\theta_{i1}, \theta_{i2}, \cdots, \theta_{iD}]^\top$ is a vector in $\mathbb{R}^D$ describing a set of $D$ latent traits,
$\bm{a}_j$ is a vector of $D$ discrimination parameters for the $j^{th}$ item, indicating the relative importance of each trait in correctly answering the item, 
and the intercept parameter $d_j$ is a scalar for item $j$. 
An applicable item $j$ takes each element of $\bm{a}_j$ to be nonnegative. 
Therefore, as each element's value of $\bm{\theta}_i$ increases, the probability of correct response increases.

\blue{With an online calibration design, an accurately calibrated item bank can be acquired using previous learners' response data for an adaptive learning system without large pretest subject pools \citep{makransky2014automatic, zhang2016smart}.
After item parameters are pre-calibrated, a variety of latent trait estimation methods can be applied to estimate learners' abilities.}
Conventional methods such as maximum likelihood estimation \citep{lord1968statistical}, weighted likelihood estimation and Bayesian methods (e.g. expected a posteriori estimation (EAP), maximum a posteriori (MAP)) can accurately estimate latent traits in MIRT models. 
Their variations are also extended for estimating the latent traits in multiple dimensions. 
Many latent trait estimation methods result in a bias on the order of as small as $O(n^{-1})$, where $n$ denotes test length, while approaches that further reduce the bias as well as the variance of estimates have also been identified and proposed \citep{firth1993bias, tseng2001multidimensional, wang2015latent, warm1989weighted, zhang2011investigating}.

\subsection{Assumptions}

Denoted $\bm{\theta}^{(t)} = [\theta_1^{(t)}, \cdots, \theta_D^{(t)}]^\top$ as learner's latent traits at time step $t$, where $D$ is the number of dimensions.
Throughout this paper, we make the following simplifying yet practical assumptions:
\begin{itemize}
	\item[{\textbf A1}.]~ No retrogression exists in latent traits. That is, $\theta_d^{(t+1)} \geq \theta_d^{(t)}$, $\forall d \in \{1, \cdots, D\}$.
	\item[{\textbf A2}.]~ The number of learning materials is finite.
\end{itemize}

%%%%%%%%%%%%%%%%%%%%%%%%%%%%%%%%%%%%%%%%%%%%%%%%%%%%%%%%%%%%%%%%%%%%%%
\section{Adaptive Learning Problem}

In this section, we first describe the adaptive learning problem and then formulate this problem as an MDP.

\subsection{Problem Statement}

\begin{figure}[H]
\centering
\includegraphics[scale=0.2]{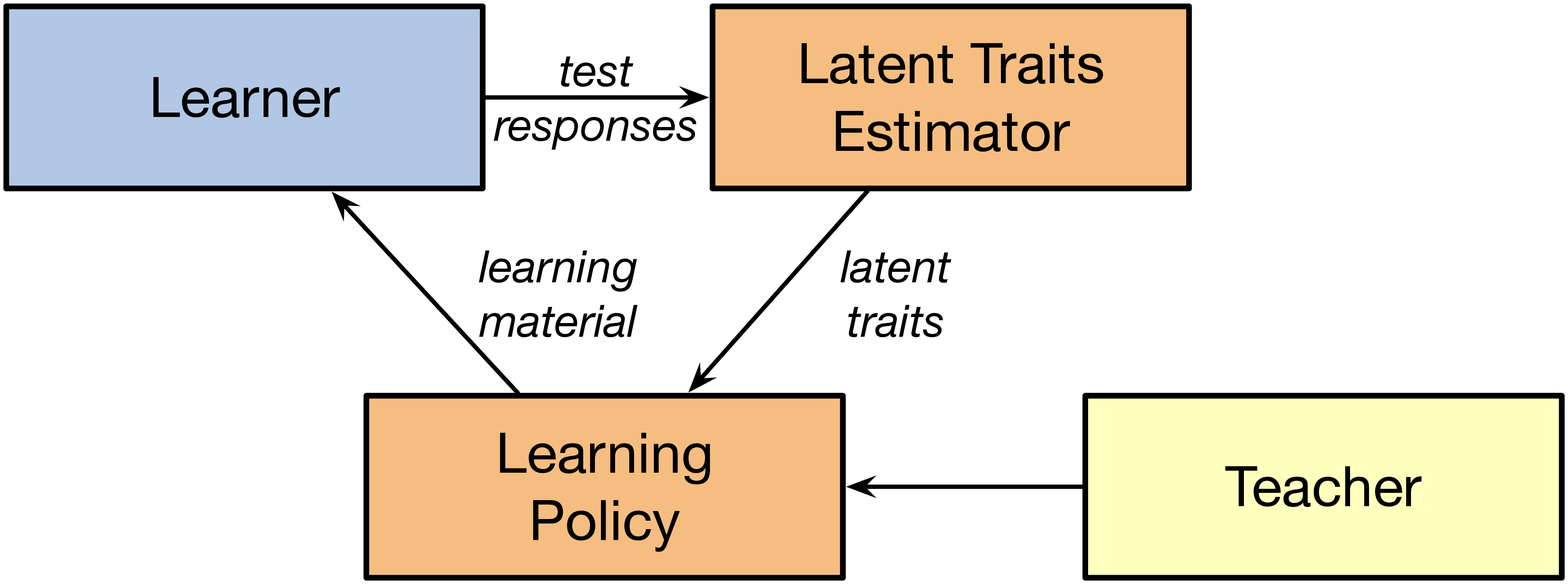}
\caption{Conventional adaptive learning system.\label{fig:e_learning_conv}}
\end{figure}

\blue{A conventional adaptive learning system is illustrated in Figure \ref{fig:e_learning_conv}.
Such an adaptive learning system is typical in traditional classrooms and online courses like Massive Open Online Courses (MOOCs) \citep{lan2016contextual}.
In the adaptive learning system, the learner takes some learning materials to improve her/his latent traits.
After the learner finishes learning the materials, a test or homework is assigned to the learner.
Then, the learner's latent traits are estimated.
Based on the estimated latent traits, the learning system adaptively determines the next learning material for the learner, which may be one of many forms including a textbook chapter, a lecture video, an interactive task, an instructor support, or an instruction pace.
Such cyclic learning process continues until the learner's latent traits reach or are close to a prespecified levels of proficiency.
}

The tests in an adaptive learning system can be computerized adaptive testing (CAT). 
The CAT is a test mode that administers tests adapted to test takers' trait levels \citep{chang2015psychometrics}.
It provides more accurate trait estimates with much smaller number of items \citep{weiss1982improving} by sequentially selecting and administering items tailored to each individual learner. 
Therefore, a relatively short test can assess learners' latent traits with high accuracy.

\blue{
Conventionally, the learning policy (or plan) is provided by a teacher as illustrated in Figure \ref{fig:e_learning_conv}. 
As aforementioned, however, it is too expensive for teachers to make an individualized adaptive learning policy for each learner.
In this paper, we use a DRL algorithm to search for an optimally individualized adaptive learning policy for each learner.
The algorithm selects the most appropriate learning material among all available materials for each learner based on her/his provisional estimated latent traits that are obtained from her/his learning history and performances in tests.
The adaptive selection of learning materials guarantees the learner reaches a prespecified proficiency level in a shortest number of learning cycles or reaches proficiency level as high as possible in a fixed number of learning cycles.
That is, instead of resorting to an experienced teacher for the construction of a learning policy as illustrated in Figure \ref{fig:e_learning_conv}, we will develop a systematic method to enable the adaptive learning system to discover an optimal learning policy from the data that have been collected, which include historical learning materials, test responses, and estimated latent traits, etc.}

\subsection{Markov Decision Process Formulation}

\subsubsection{Primer on Markov Decision Process}

Before presenting the formulation for the adaptive learning problem, we first briefly review MDPs.
\blue{An MDP is characterized by a 5-tuple $(\mathcal{S}, \mathcal{A}, \mathcal{P}, \mathcal{R}, \mathcal \gamma)$, where $\mathcal{S}$ is a set of states, $\mathcal{A}$ is a set of actions, $\mathcal{P}$ is a Markovian transition model, $\mathcal{R}: \mathcal{S} \times \mathcal{A} \times \mathcal{S} \rightarrow \mathbb R$ is a reward function, and $\gamma \in [0,1)$ is a discount factor \citep{sutton2018reinforcement}. }
A transition sample is defined as $(\bm{s},\bm{a},r,\bm{s}')$, where $\bm{s}, \bm{s}' \in \mathcal{S}$ and $\bm{a} \in \mathcal{A}$, $r=\mathcal{R}(\bm{s},\bm{a},\bm{s}')$ is a scalar reward when the state transitions into state $\bm{s}'$ from state $\bm{s}$ after taking action $\bm{a}$.

Let $\bm{S}^{(t)}$ and $\bm{A}^{(t)}$ denote the state and action at time step $t$, respectively, and $R^{(t)}$ denote the reward obtained after taking action $\bm{A}^{(t)}$ at state $\bm{S}^{(t)}$. 
Note that $\bm{S}^{(t)}$, $\bm{A}^{(t)}$, and $R^{(t)}$ are random variables.
When both $\mathcal{S}$ and $\mathcal{A}$ are finite, the transition model $\mathcal{P}$ can be represented by conditional probability, that is, 
\begin{align} \label{eq5}
	\mathcal{P}^{(t)}(\bm{s}'|\bm{s},\bm{a}) = \mathbb{P}(\bm{S}^{(t+1)}=\bm{s}'|\bm{S}^{(t)}=\bm{s},\bm{A}^{(t)}=\bm{a}).
\end{align} 
The Markovian property of the transition model is that, for any time step $t$, 
\begin{align} \label{eq3}
	\mathbb{P}(\bm{S}^{(t+1)}|\bm{A}^{(t)},\bm{S}^{(t)},\dots,\bm{A}^{(0)},\bm{S}^{(0)})=\mathbb{P}(\bm{S}^{(t+1)}|\bm{A}^{(t)},\bm{S}^{(t)}).
\end{align}
Essentially, the Markovian property requires that a future state is independent of all past states given the current state. 
Assume $\mathcal{P}$ is time-homogeneous, i.e., for any two time steps $t_1$ and $t_2$,
\begin{align} \label{eq7}
	\mathcal{P}^{(t_1)}(\bm{s}'|\bm{s},\bm{a}) = \mathcal{P}^{(t_2)}(\bm{s}'|\bm{s},\bm{a}).
\end{align}
Then, we can drop the superscript $t$ and write the transition model as $\mathcal{P}(\bm{s}'|\bm{s},\bm{a})$.
Note that when $\mathcal{S}$ is continuous, the transition model can be represented by a conditional probability density function.

%Let $\Delta \bm{S}^{(t)}$ denote the change of state from time step $t$ to time $t+1$, that is $\Delta \bm{S}^{(t)}=\bm{S}^{(t+1)}-\bm{S}^{(t)}$, and $\bm{\Delta s}=\bm{s}'-\bm{s}$, then the probability in equation \eqref{eq5} can also be represented as
%\begin{equation} \label{eq6}
%	\mathbb{P}(\bm{S}^{(t+1)}=\bm{s}'|\bm{S}^{(t)}=\bm{s},\bm{A}^{(t)}=\bm{a})=\mathbb{P}(\Delta \bm{S}^{(t)}=\Delta \bm{s}|\bm{S}^{(t)}=\bm{s},\bm{A}^{(t)}=\bm{a}).
%\end{equation}

\blue{Let $\pi:\mathcal{S} \rightarrow \mathcal{A}$ denote a deterministic policy for the MDP defined above.
The action-value function for the MDP under policy $\pi$ is defined as follows:
\begin{align}
	Q^{\pi}(\bm{s}, \bm{a}) = \mathbb{E}[\sum_{t=0}^\infty \gamma^t R^{(t)} | \bm{S}^{(0)} = \bm{s}, \bm{A}^{(0)} = \bm{a}; \pi],
\end{align}
where $\mathbb{E}$ denotes the expectation.
The action-value function $Q^{\pi}(\bm{s}, \bm{a})$ is the expected cumulative discounted reward when the system starts from state $\bm{s}$, takes action $\bm{a}$, and follows policy $\pi$ thereafter.
The maximum action-value function over all policies is defined as $Q(\bm{s}, \bm{a}) = \max_{\pi} Q^{\pi}(\bm{s}, \bm{a})$.
A policy $\pi$ is said to be optimal if $Q^{\pi}(\bm{s}, \bm{a}) = Q(\bm{s}, \bm{a})$ for any $\bm{s} \in \mathcal{S}$ and $\bm{a} \in \mathcal{A}$.
In particular, the greedy policy with respect to $Q(\bm{s}, \bm{a})$, defined as $\pi^*(\bm{s}) = \argmax_{\bm{a}} Q(\bm{s}, \bm{a})$, is an optimal policy \citep{sutton2018reinforcement}.
The MDP is solved if we find $\pi^*$.}
\begin{theorem}\citep{bertsekas1996neuro}
The optimal action-value function $Q(\bm{s}, \bm{a})$ satisfies the Bellman optimality equation:
\begin{align} \label{eq:bellman_opt}
	Q(\bm{s}, \bm{a}) = \mathbb{E}[R^{(0)}] + \gamma \sum \limits_{\bm{s}' \in \mathcal{S}} \mathcal{P}(\bm{s}' | \bm{s}, \bm{a}) \max_{\bm{a}' \in \mathcal{A}} Q(\bm{s}', \bm{a}').
\end{align}
Furthermore, there is only one $Q$ function that solves the Bellman optimality equation.
\end{theorem}
%\begin{align} \label{eq:bellman_opt}
%	Q(\bm{s}, \bm{a}) = \mathbb{E}[R^{(0)} + \gamma \max_{\bm{a}' \in \mathcal{A}} Q(\bm{S}', \bm{a}') | \bm{s}, \bm{a}].
%\end{align}
The Bellman optimality equation is of central importance to solving the MDP.
When both $\mathcal{S}$ and $\mathcal{A}$ are finite and $\mathcal{P}$ is known, model-based based algorithms such as the value iteration algorithm can be applied to solve the MDP \citep{sutton2018reinforcement}.

\subsubsection{Adaptive Learning Problem as MDP}

We next formulate the adaptive learning problem as an MDP as follows.

\noindent \textit{State Space:} Define the vector of parameters describing the learner's latent traits as the state, i.e., $\bm{s} = \bm{\theta}$, which has $D$ continuous variables, where $D$ represents the dimension of the latent traits. 
\blue{
For the simplicity of the algorithm construction in the following, the state space is defined as $\mathcal{S} = [0, 1]^D$ when each element of $\bm{\theta}$ satisfies $\theta \in [0, 1]$, in which a smaller value of $\theta$ indicates a lower ability and a larger value indicates a higher ability.
Although a latent trait variable is typically defined on $\mathbb R$ in IRT, a closed interval, say $[-5, 5]$, is used as the range of a latent trait variable in practice.
Let $h_d$ be the prespecified target proficiency level of the $d^{\text{th}}$ latent trait, which is the level the learners try to reach, where $d=1, \dots, D$.
Because of the fact that there is a bijection between $[-5, h_d]$ and $[0, 1]$, an estimated trait $\theta \in [-5, h_d]$ can be directly transformed into the scale of $[0, 1]$.
Thus, without loss of generality, we consider the state space as $\mathcal{S} = [0, 1]^D$.
}

\noindent \textit{Action Space:} Let the learning materials available in the adaptive learning system be indexed by $1, 2, \cdots, L$. The action $\bm{a}$ in the adaptive learning system is the index of a learning material, which is discrete, and the action space is $\mathcal{A} = \{1 , \cdots, L\}$.
%\footnote{Note that since the action is a scaler, we use $a$ (not in bold) instead of $\bm{a}$ (in bold) to denote the action throughout the rest of the paper.}

\noindent \textit{Reward Function:} Recall that the objective of the adaptive learning system is to minimize the learning steps it takes before a learner's latent traits reach the maximum, i.e., for $\bm{\theta}$ to reach $\bm{1}_D$, where $\bm{1}_D$ is an all-ones vector in $\mathbb{R}^D$. As such, the reward function is defined as follows:
	\begin{align}
		r = \mathcal{R}(\bm{s},\bm{a},\bm{s}') = \left\{ \begin{array}{ll}
		-1, & \text{if}~||\bm{s}' - \bm{1}_D||_\infty \geq 10^{-3}, \\
		0, & \textrm{otherwise},
		\end{array} \right.
	\end{align}
	where $||\cdot||_\infty$ indicates the infinite norm.
	Intuitively, the sum of rewards over one episode (\blue{the entire learning process of a learner}) is to the negative of the total steps a learner takes before all of her/his latent traits \blue{are very close to $1$, which indicates that the learner has reached target levels of all prespecified abilities}.

\noindent \textit{Transition Model:} 
The probability distributions of the latent trait as well as the change of trait are unknown. As a result, the transition model $\mathcal{P}(\bm{s}'|\bm{s},\bm{a})$ is not known a priori. 

\blue{Based on this MDP formulation, the adaptive learning problem is essentially to find an optimal learning policy, denoted by $\pi^*:\mathcal{S} \rightarrow \mathcal{A}$, that determines the action (learning material selection) based on the state (latent traits), such that the expected cumulative discounted reward is maximized.}
Note that the larger the expected cumulative discounted reward is, the less the total learning steps a learner takes to reach the target level(s) of an ability/abilities is.
\blue{Since the transition model $\mathcal{P}$ is unknown, the MDP cannot be solved using model-based algorithms such as the value iteration algorithm.}
We will resort to a model-free DRL algorithm to solve it in the next section.

%%%%%%%%%%%%%%%%%%%%%%%%%%%%%%%%%%%%%%%%%%%%%%%%%%%%%%%%%%%%%%%%%%%%%%
\section{Optimal Learning Policy Discovery Algorithm}

In this section, we solve the adaptive learning problem by using the deep Q-learning algorithm, which can learn the action-value function directly from historical transition data without knowing the underlying transition model.
To utilize the available transition information more efficiently, we further develop a transition model estimator and use it to train the deep Q-learning algorithm.

\subsection{Action-Value Function As Deep Q-Network}

Recall that the optimal learning policy can be readily obtained if we know the action-value function.
When the state is continuous and the action is discrete, which is the case in the adaptive learning problem, the action-value function $Q(\bm{s}, \bm{a})$ cannot be exactly represented in a tabular form.
\blue{In such cases, the action-value function can be approximated by some functions, such as linear functions \citep{sutton2018reinforcement} or artificial neural networks (simply referred to as neural networks) \citep{mnih2015human}.
In the former case, the approximate action-value function is represented as an inner product of the parameter vector and a feature vector that is constructed from the state.
It is important to point out the choice of the features is critical to the performance of the approximate action-value function.
Meanwhile, neural networks are capable of extracting useful features from the state directly, and have stronger representation power than linear functions \citep{goodfellow2016deep}.}

\begin{figure}[H]
\centering
\includegraphics[scale=0.2]{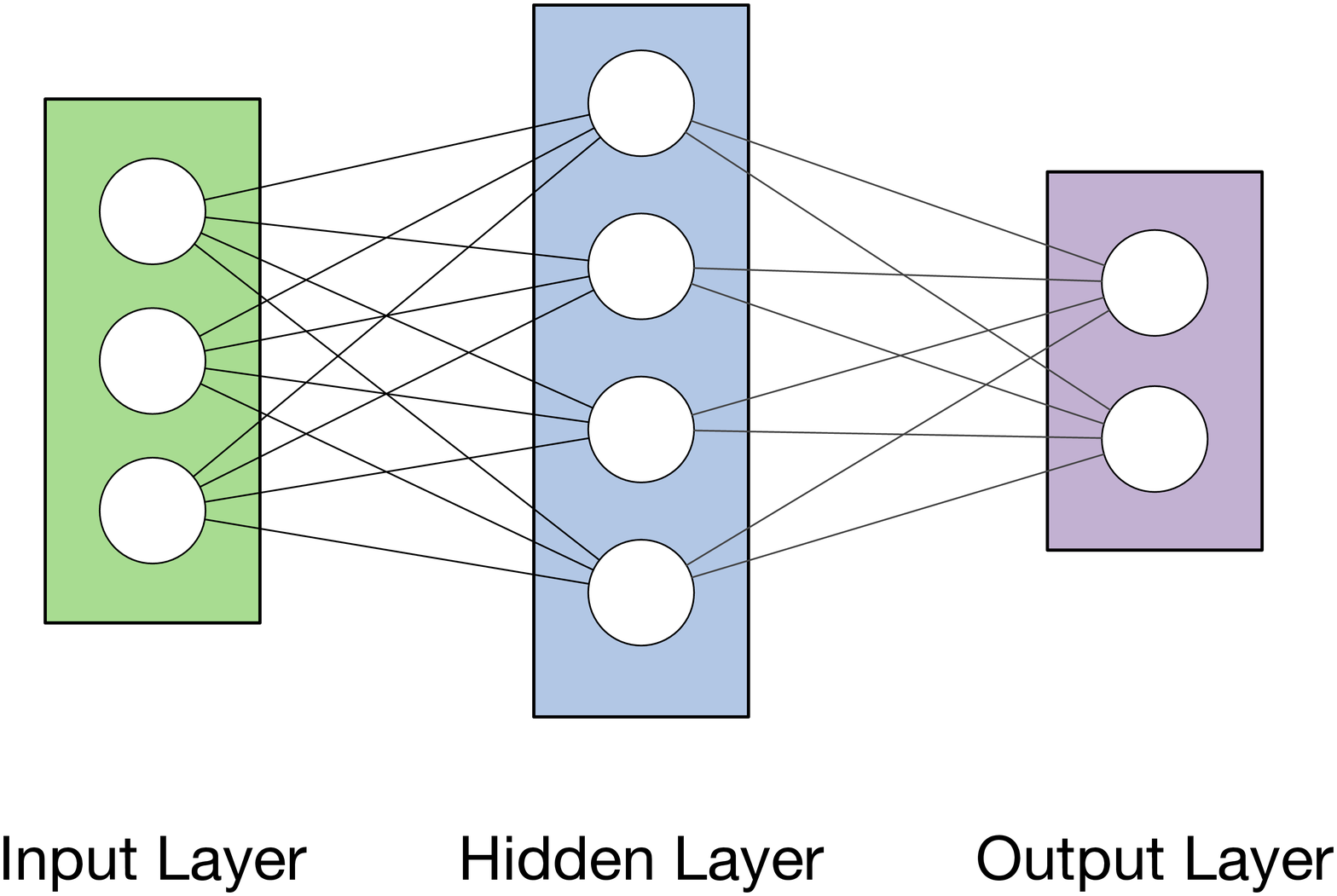}
\caption{An illustrative neural network with one hidden layer.\label{fig:nn}}
\end{figure}

As an example for neural networks, Fig. \ref{fig:nn} shows an illustrative neural network that consists of an input layer that has $3$ units, a hidden layer that has $4$ units, and an output layer with $2$ units.
Let $\bm{x} = [x_1, x_2, x_3]^\top$, $\bm{h} = [h_1, h_2, h_3, h_4]^\top$, and $\bm{y} = [y_1, y_2]^\top$ denote the vectors that come out of the input layer, the hidden layer, and the output layer, respectively.
In the neural network, the output of one layer is the input for the next layer.
To be more specific, $\bm{h}$ can be computed from $\bm{x}$, and $\bm{y}$ can be computed from $\bm{h}$ as follows:
\begin{align}
	\bm{h} &= \phi(\bm{W}_{hx} \bm{x} + \bm{b}_h), \\
	\bm{y} &= \bm{W}_{yh} \bm{h} + \bm{b}_y,
\end{align}
where $\bm{W}_{hx} \in \mathbb{R}^{4 \times 3}$ and $\bm{W}_{yh} \in \mathbb{R}^{2 \times 4}$ are two weight matrices, $\bm{b}_h \in \mathbb{R}^{4}$ and $\bm{b}_y \in \mathbb{R}^{2}$ are two bias vectors, and $\phi(\cdot)$ is the so-called activation function, which is applied to its argument element-wise.
A popular choice of the activation function $\phi$ is the rectifier, i.e., $\phi(x) = \max(x, 0)$.
Conceptually, we can write the output $\bm{y}$ as a function of $\bm{y} = \varphi(\bm{x})$, where $\varphi(\cdot)$ is parameterized by $\bm{W}_{hx}$, $\bm{W}_{yh}$, $\bm{b}_h$, and $\bm{b}_y$, which can be collectively denoted as a parameter vector $\bm{w}$.
Given a set of input-output values denoted by $\{(\bm{x}^{(i)}, \bm{y}^{(i)}): i = 1, \cdots, M \}$, the optimal value of $\bm{w}$ can be found by solving the following problem:
\begin{align} \label{eq:nn_min}
	\min_{\bm{w}} \sum_{i=1}^M || \varphi(\bm{x}^{(i)}) - \bm{y}^{(i)} ||^2,
\end{align}
where $||\cdot||$ is the $L_2$-norm.
Problem \eqref{eq:nn_min} can be solved by using gradient descent algorithm or its variants, in which the gradient of the objective function with respect to $\bm{w}$ can be computed using the famous backpropagation technique. 
Neural networks can also be trained using a variety of other optimization algorithms such as Adam and RMSProp \citep[see,][]{goodfellow2016deep}.
Note that there may be several hidden layers and the more hidden layers there are, the deeper the neural network is.
We refer interested readers to \citet{goodfellow2016deep} for a more comprehensive details about neural networks.

Recall that in the adaptive learning problem, the state is continuous in $[0, 1]^D$, while the action is discrete $\mathcal{A} = \{1, \cdots, L\}$.
The approximate action-value function, denoted by $\hat{Q}(\bm{s}, \bm{a})$, can be represented using a neural network as follows.
The input layer is the state $\bm{s}$, or equivalently, the latent trait vector $\bm{\theta}$, which has $D$ units.
The output has $L$ units, each of which corresponds to the action-value for one action.
To more be specific, the $\ell^{\text{th}}$ unit in the output layer gives $\hat{Q}(\bm{s}, \bm{a}=\ell)$, i.e., the action-value for state $\bm{s}$ and action $\ell$.
The number of hidden layers and the number of units in each hidden layer can be determined through simulation, which is to be detailed in the numerical simulation section.
Such a neural network is also referred to as a deep Q-network (DQN) \citep{mnih2013playing}.
Let $\bm{w}$ denote the parameter vector of the DQN, which includes all weights and biases in the DQN.
To emphasize that $\hat{Q}(\bm{s}, \bm{a})$ is parameterized by $\bm{w}$, we write $\hat{Q}(\bm{s}, \bm{a})$ as $\hat{Q}(\bm{s}, \bm{a}; \bm{w})$.

\begin{figure}[H]
\centering
\includegraphics[scale=0.2]{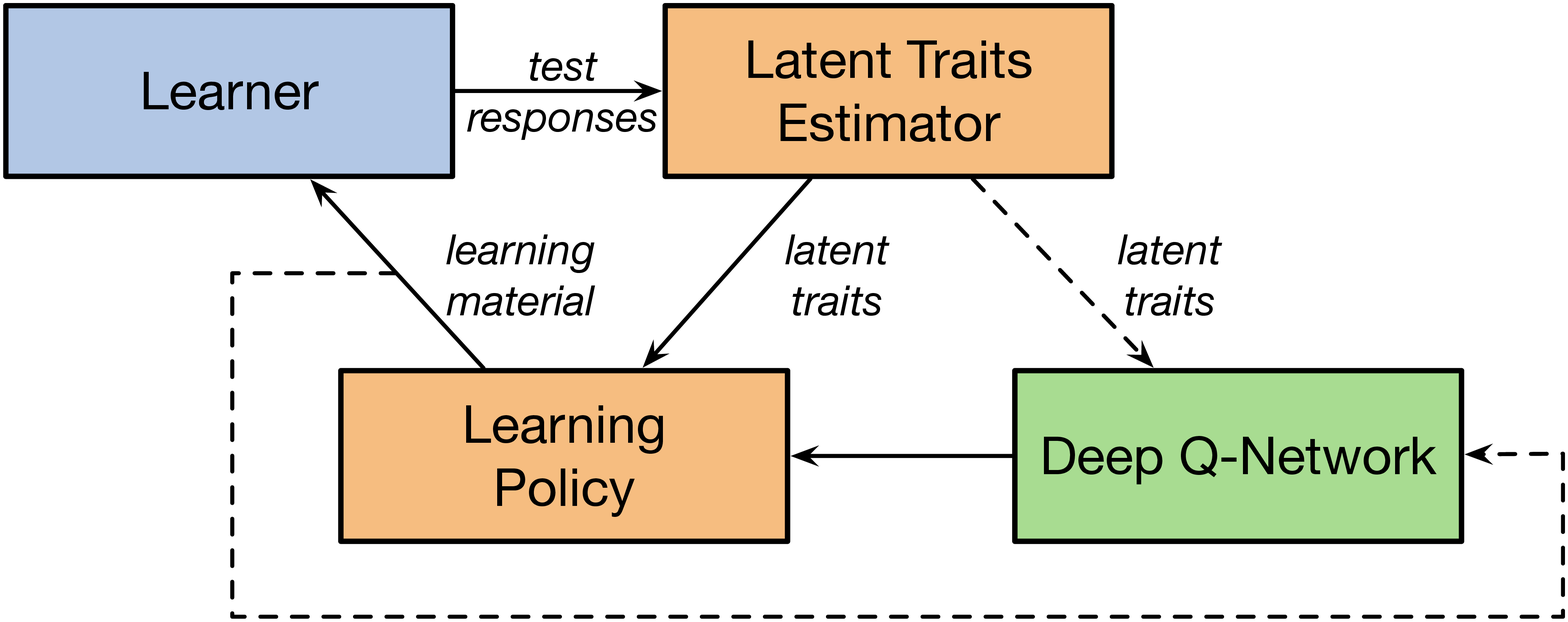}
\caption{Adaptive adaptive learning system with DQN.\label{fig:e_learning_dqn}}
\end{figure}

Once we have $\hat{Q}(\bm{s}, \bm{a}; \bm{w})$, the optimal learning policy becomes readily available, which is $\pi^*(\bm{s}) = \argmax_{\bm{a}} \hat{Q}(\bm{s}, \bm{a}; \bm{w})$.
Then, the ``Teacher" block in Figure \ref{fig:e_learning_conv} can be replaced with the DQN as shown in Figure \ref{fig:e_learning_dqn}.

\subsection{Learning Policy Discovery with Deep Q-Learning}

The parameters of the DQN can be learned from the the sequence of latent traits and learning materials using the deep Q-learning algorithm proposed by \citet{mnih2013playing}.
The optimal value of the parameter vector of the DQN, $\bm{w}$, can be found by minimizing the mean squared error between the approximate action-value function and the true action-value function:
\begin{align} \label{eq:dqn_min}
	\min_{\bm{w}} \mathbb{E}[( \hat{Q}(\bm{S}, \bm{A}; \bm{w}) - Q(\bm{S}, \bm{A}) )^2].
\end{align}
However, solving \eqref{eq:dqn_min} is extremely difficult if not impossible since both $Q(\bm{S}, \bm{A})$ and the transition model are unknown and thus, the expectation of the mean squared error cannot be computed.
The deep Q-learning algorithm adopts two measures to cope with these challenges.
First, the expectation is replaced with the sample average that can be computed from a set of historical transitions, denote by $\mathcal{M} = \{(\bm{s}, \bm{a}, r, \bm{s}'): \bm{s}, \bm{s}' \in \mathcal{S}, \bm{a} \in \mathcal{A} \}$, with $|\mathcal{M}| = M$, where $|\cdot|$ denotes the cardinality of a set.
That is, \eqref{eq:dqn_min} is now replaced by the following problem:
\begin{align} \label{eq:dqn_min_sample}
	\min_{\bm{w}} \sum_{(\bm{s}, \bm{a}, r, \bm{s}') \in \mathcal{M}}( \hat{Q}(\bm{s}, \bm{a}; \bm{w}) - Q(\bm{s}, \bm{a}) )^2.
\end{align}
At time step $t$, the parameter vector is updated using the gradient descent algorithm as follows:
\begin{align} \label{eq:dqn_gd}
	\bm{w}^{(t+1)} = \bm{w}^{(t)} - \alpha \sum_{(\bm{s}, \bm{a}, r, \bm{s}') \in \mathcal{M}}( \hat{Q}(\bm{s}, \bm{a}; \bm{w}^{(t)}) - Q(\bm{s}, \bm{a}) ) \frac{\partial \hat{Q}(\bm{s}, \bm{a}; \bm{w}^{(t)})}{\partial \bm{w}},
\end{align}
where $\alpha > 0$ is the learning rate and $\bm{w}^{(t)}$ denotes the value of $\bm{w}$ at time step $\bm{w}$.
Second, the unknown $Q(\bm{s}, \bm{a})$ is further substituted by $r + \gamma \max_{\bm{a}'} \hat{Q}(\bm{s}', \bm{a}'; \bm{w}^{(t)})$ based on the Bellman optimality equation in \eqref{eq:bellman_opt}.
Note that when $||\bm{s}' - \bm{1}_D||_\infty < 10^{-3}$, which indicates the learning process has ended, $Q(\bm{s}', \bm{a}') = 0$.
Therefore, \eqref{eq:dqn_gd} is now becomes
\begin{align} \label{eq:dqn_gd2}
	\bm{w}^{(t+1)} = \bm{w}^{(t)} - \alpha \sum_{(\bm{s}, \bm{a}, r, \bm{s}') \in \mathcal{M}}( \hat{Q}(\bm{s}, \bm{a}; \bm{w}^{(t)}) - y) \frac{\partial \hat{Q}(\bm{s}, \bm{a}; \bm{w}^{(t)})}{\partial \bm{w}},
\end{align}
where 
\begin{align}
	y = \left\{ \begin{array}{ll}
		r, & \text{if}~||\bm{s}' - \bm{1}_D||_\infty < 10^{-3}, \\
		r + \gamma \max_{\bm{a}'} \hat{Q}(\bm{s}', \bm{a}'; \bm{w}^{(t)}), & \textrm{otherwise}.
		\end{array} \right.
\end{align}

The detailed deep Q-learning algorithm that is used to search the optimal parameter vector for the DQN is presented in Algorithm \ref{algo:dqn}, where \blue{one episode represents a complete learning process of one learner and the number of episodes is the number of learners}.
In order to obtain a good approximate action-value function, the state-action space needs to be sufficiently explored.
To achieve this, the so-called $\epsilon$-greedy exploration is adopted in the deep Q-learning algorithm.
Specifically, at time step $t$, a random action $\bm{a}^{(t)}$ is selected with probability $\epsilon^{(t)}$, and a greedy action $\bm{a}^{(t)} = \max_{\bm{a}} \hat{Q}(\bm{s}^{(t)}, \bm{a}; \bm{w}^{(t)})$ is with probability $1 - \epsilon^{(t)}$.
In this paper, we adaptively decay $\epsilon^{(t)}$ from $\overline{\epsilon}$ to $\underline{\epsilon}$ in $\tau_\epsilon$ time steps.
In addition, the parameter vector is updated at each time step using a set of transitions $\mathcal{M}$ that is resampled from the historical transitions denoted by $\mathcal{H}$ with $|\mathcal{H}| = H$ so as to reduce the bias that may be caused by the samples.

\begin{algorithm}[!t]
    \SetAlgoLined
    \DontPrintSemicolon
    \KwData{$\gamma, \alpha, \overline{\epsilon}, \underline{\epsilon}, \tau_\epsilon, M, E$}
    \KwResult{$\bm{w}$}
	    Randomly initialize $\bm{w}$ and set total time step counter $\tau = 0$\;
	    \For{episode $=1, \cdots, E$ }{
	    	Receive initial state $\bm{s}_{0}$\;
		    \For{$t = 0, 1, \cdots, $}{
		    	Compute $\epsilon^{(t)} = \overline{\epsilon} - (\overline{\epsilon} - \underline{\epsilon}) \times \min(\tau/\tau_\epsilon, 1)$ and increase $\tau$ by $1$\;
		    	With probability $\epsilon^{(t)}$ select a random action $\bm{a}^{(t)}$ otherwise select $\bm{a}^{(t)} = \max_{\bm{a}} \hat{Q}(\bm{s}^{(t)}, \bm{a}; \bm{w}^{(t)})$\;
		    	Send the learning material determined by $\bm{a}^{(t)}$ to the learner\;
		    	Given the learner a test and collect test response\;
		    	Receive new state $\bm{s}^{(t+1)}$ estimated from test response by latent trait estimator\;
		        Compute reward $r^{(t)}$ according to
		        \begin{align*}
					r^{(t)} = \left\{ \begin{array}{ll}
					-1, & \text{if}~||\bm{s}^{(t+1)} - \bm{1}_D||_\infty \geq 10^{-3} \\
					0, & \textrm{otherwise}
					\end{array} \right.
				\end{align*}\;\vspace{-0.5in}
		    	Store transition $(\bm{s}^{(t)}, \bm{a}^{(t)}, r^{(t)}, \bm{s}^{(t+1)})$ into $\mathcal{H}$\;
		    	Sample $M$ transitions from $\mathcal{H}$ and store them into $\mathcal{M}$\;
		    	Update $\bm{w}$ according to
				\begin{align*}
					\bm{w}^{(t+1)} = \bm{w}^{(t)} - \alpha \sum_{(\bm{s}, \bm{a}, r, \bm{s}') \in \mathcal{M}}( \hat{Q}(\bm{s}, \bm{a}; \bm{w}^{(t)}) - y) \frac{\partial \hat{Q}(\bm{s}, \bm{a}; \bm{w}^{(t)})}{\partial \bm{w}}
				\end{align*}
				where 
				\begin{align*}
					y = \left\{ \begin{array}{ll}
						r, & \text{if}~||\bm{s}' - \bm{1}_D||_\infty < 10^{-3} \\
						r + \gamma \max_{\bm{a}'} \hat{Q}(\bm{s}', \bm{a}'; \bm{w}^{(t)}), & \textrm{otherwise}
						\end{array} \right.
				\end{align*}\;\vspace{-0.5in}
				\textbf{if} $||\bm{s}' - \bm{1}_D||_\infty < 10^{-3}$, break\;
		    }
	    }
\caption{Deep Q Learning Algorithm for Adaptive Learning Problem}
\label{algo:dqn}
\end{algorithm}

\subsection{Transition Model Estimator}

\blue{The deep Q-learning algorithm requires a sufficiently large historical transition data in order to find a good approximate of the action-value function, based on which the learning policy is then derived.
However, we may not be able to obtain adequate transitions due to several reasons including the lack of adequate learners, and the long time it takes to acquire an individual learner's learning path (transitions). 
Thus, it is more desirable to develop an adaptive learning system which can efficiently discover the optimal learning policy after training on a relatively small number of learners. 
To this end, we develop a transition model estimator which emulates the learning behavior of learners.}
Specifically, the transition model estimator can take a state $\bm{s}$ and an action $\bm{a}$ as inputs, and output the next state $\bm{s}'$.
This can be cast as a supervised learning task, (a regression task), which can be solved using neural networks.
The input layer of the neural network that represents the transition model is a pair of state and action, and the output layer is the next state.
The number of hidden layers can be adjusted through the parameter tuning process \citep[see, e.g.,][for more details]{goodfellow2016deep}.

Conceptually, we can write the neural network that represents the transition model as $\bm{s}' = \psi(\bm{s}, \bm{a})$, the parameter vector of which is denoted by $\bm{v}$.
The optimal value of $\bm{v}$ can be found by solving the following problem using the backpropagation algorithm:
\begin{align}
	\min_{\bm{v}} \sum_{(\bm{s}, \bm{a}, r, \bm{s}') \in \mathcal{H}} || \psi(\bm{s}, \bm{a}) - \bm{s}' ||^2,
\end{align}
where $\mathcal{H}$ is the set of historical transition (data).

\begin{figure}[H]
\centering
\includegraphics[scale=0.2]{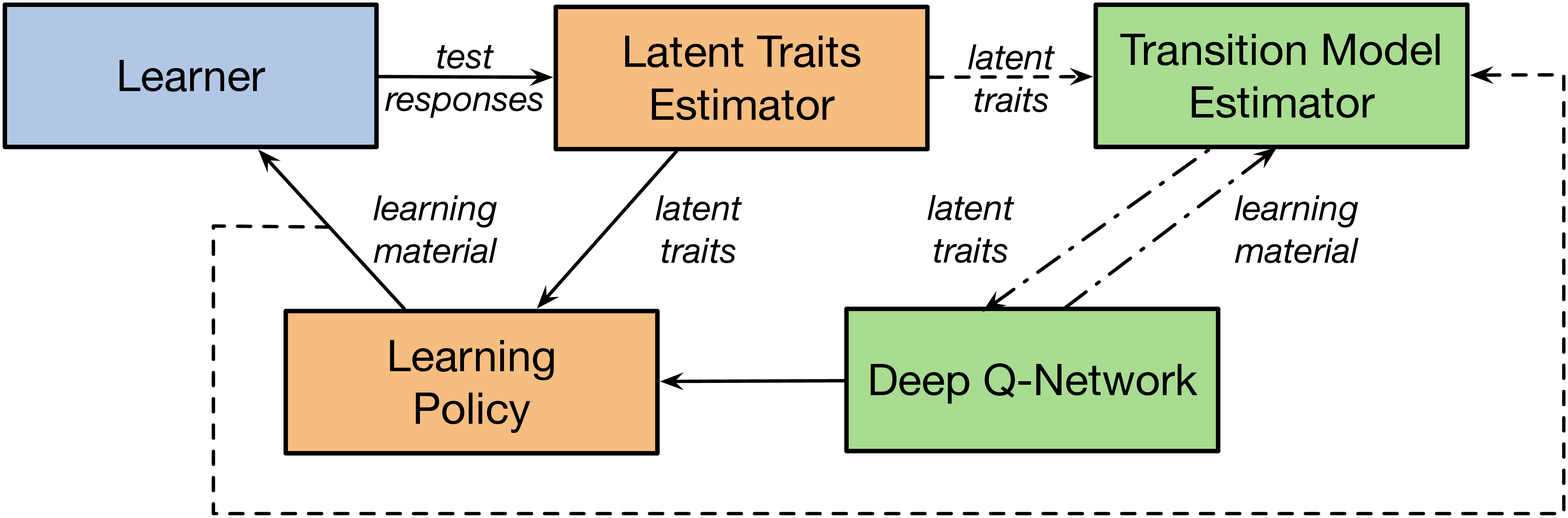}
\caption{Adaptive adaptive learning system with DQN and transition model estimator.\label{fig:e_learning_dqn_nn}}
\end{figure}

The adaptive learning system with the DQN and a transition model estimator is shown in Fig. \ref{fig:e_learning_dqn_nn}, where the DQN is trained against the transition model, instead of the actual learners.

%%%%%%%%%%%%%%%%%%%%%%%%%%%%%%%%%%%%%%%%%%%%%%%%%%%%%%%%%%%%%%%%%%%%%%
\section{Numerical Simulation}

In this section, we show the performance of the adaptive learning system with and without the transition model estimator, and also investigate the impacts of latent trait estimation errors through two simulation studies.

\subsection{Simulation Overview}

\blue{Consider a group of learners in a two-dimensional assessment and a learning environment with three sets of learning materials.  
We model the group of learners as a homogeneous MDP.}
Let the random vector $\bm{\Theta}^{(t)} = [\Theta_1^{(t)}, \Theta_2^{(t)}]^\top$ denote a learner's state $\bm{S}^{(t)}$ at time step $t$, which represents the latent traits in our study.
\blue{
Consider three sets of learning materials regarding the two-dimensional latent trait levels, that is, $\mathcal{A} = \{1, 2, 3\}$. 
Each set of learning materials contain contents with regards to different latent traits.}
Denote the change of the latent traits from time step $t$ to $t+1$ by $\Delta \bm{\Theta}^{(t)} = [\Delta \Theta_1^{(t)}, \Delta \Theta_2^{(t)}]^\top$.
The probability of having $\Delta \bm{\theta} = [\Delta \theta_1,\Delta \theta_2]^\top$ transitioning from state $\bm{\theta} = [\theta_1,\theta_2]^\top$ to $\bm{\theta'} = [\theta_1',\theta_2']^\top$ can be represented as
\begin{align} \label{eq8}
	\mathcal{P}(\bm{\theta'} | \bm{\theta}, \bm{a}) = \mathbb{P}(\Delta \bm{\Theta}^{(t)} = \bm{\Delta \theta} | \bm{\Theta}^{(t)} = \bm{\theta}, \bm{A}^{(t)} = \bm{a}),
\end{align}
\blue{where $\bm{a}$ is the index of the set which the selected learning material belongs to. 
In the following notations, we only consider the set which the selected learning material belongs to, denoted as $\bm{a}$.}
Assume $\theta_1,\theta_2 \in [0,1]$, where the value of $0$ indicates extremely low ability on the corresponding dimension and the value of $1$ indicates the target ability. 
%The conditional probability density function of $\Delta \bm{\Theta}$ given the occurrence of the value $\bm{\theta}$ and $a$ of $\bm{\Theta}^{(t)}$ and $\bm{A}^{(t)}$ respectively can be written as
%\begin{align} \label{eq9}
%\begin{gathered}
%	f(\Delta \theta_1, \Delta \theta_2|\theta_1,\theta_2,a)
%	=\dfrac{f(\Delta \theta_1, \Delta \theta_2,\theta_1,\theta_2,a)}{f(\theta_1,\theta_2,a)} \\
%	=\dfrac{f(\Delta \theta_2|\Delta \theta_1,\theta_1,\theta_2,a)
%	f(\Delta \theta_1,\theta_1,\theta_2,a)}{f(\theta_1,\theta_2,a)} \\
%	=f(\Delta \theta_2|\Delta \theta_1,\theta_1,\theta_2,a)f(\Delta \theta_1|\theta_1,\theta_2,a).
%\end{gathered}
%\end{align}
%Hence, the conditional probability density function of $\Delta \bm{\Theta}$ can be represented by the product of the conditional probability density functions of $\Delta \theta_1$ and $\Delta \theta_2$. 

In addition, under Assumption \textbf{A}1 of no retrogression, we have $\Delta \theta_1 \in [0,1-\theta_1]$ and $\Delta \theta_2 \in [0,1-\theta_2]$. 
As we model the transition of the latent traits to be a continuous-state MDP, the change of $\Delta \theta_1$ and $\Delta \theta_2$ only depends on current latent trait $\bm{\theta}$ and the selected learning material $\bm{a}$. 
Therefore, we let $\Delta \theta_1$ and $\Delta \theta_2$ follow Beta distributions such that $\Delta \theta_1 \sim Beta(1, g_1(\bm{\theta}, \bm{a}))$, where $\bm{a} \in \{1,3\}$, and $\Delta \theta_2 \sim Beta(1, g_2(\Delta \theta_1, \bm{\theta}, \bm{a}))$, where $\bm{a} \in \{2,3\}$. 
\blue{$\Delta \theta_2=0$ when $\bm{a} = 1$ and $\Delta \theta_1 = 0$ when $\bm{a} = 2$, which means the first set of materials only helps improving $\theta_1$ while the second set is only related to $\theta_2$.}
Parameters of $g_1(\bm{\theta}, \bm{a})$ and $g_2(\Delta \theta_1, \bm{\theta}, \bm{a})$ in the Beta distribution are calculated by
\begin{align}\label{eq9}
g_1(\bm{\theta}, \bm{a})=\left\{
\begin{array}{ll}
3 + 8 \theta_1 - 0.2 \theta_2, & \bm{a} = 1\\
15 + 15 \theta_1 - 0.4 \theta_2,& \bm{a} = 3
\end{array} \right.
\end{align}
and
\begin{align}\label{eq10}
	g_2(\bm{\theta}, \bm{a}) = \left\{
	\begin{array}{ll}
		10 - \theta_1 + 5 \theta_2, & \bm{a} = 2 \\
		20 - 28 \theta_1 e^{-\frac{(\theta_1-0.6)^2}{0.3}} + 30 \theta_2 - 0.3 \Delta \theta_1, & \bm{a} = 3.
	\end{array} \right.
\end{align}

%where \red{$g_1(\bm{\theta},a)=\bm{w}_{1a}^\top(\bm{\theta}+\bm{1})$ and 
%$g_2(\Delta \theta_1,\bm{\theta},a)=\bm{w}_{2a}^\top (\bm{\theta}+\bm{1})+b\Delta \theta_1$. 
%Elements in vectors $\bm{w}_{1a}$ and $\bm{w}_{2a}$, $a \in \{1,2,3\}$, are presented in Table \ref{T1}. 
%As a result, values of $g_1(\bm{\theta},a)$ and $g_2(\Delta \theta_1,\bm{\theta},a)$ determine shapes of two Beta distributions.} The next state $\bm{\theta'}$ is calculated by $\theta_1'=min\{1,\theta_1+\Delta\theta_1\}$ and $\theta_2'=min\{1,\theta_2+\Delta\theta_2\}$.

%\begin{table}[!t]
%\centering
%\caption{\red{$\bm{w}_{1a}$ and $\bm{w}_{2a}$} \label{T1}}
%\begin{tabular}{lccc}
%\toprule
%Slope parameter & $\theta_1$ & $\theta_2$ & $\Delta \theta_1$ \\
%\midrule
%$\bm{w}_{11}$ & 2& 1& -\\
%$\bm{w}_{12}$& 3& 1& -\\
%$\bm{w}_{13}$& 0& 0& -\\
%$\bm{w}_{21}$ & 1& 3& -0.5\\
%$\bm{w}_{22}$& 0& 0& 0\\
%$\bm{w}_{23}$& 1& 4& 0\\
%\bottomrule
%\end{tabular}
%\end{table}

An intuitive example is how a learner learns addition ``+'' and subtraction ``--''. 
A learning process usually takes a long time and thus a monotonic decreasing, zero-concentrated distribution is adopted to simulate the ability increase. 
In that case, each learning step will most likely lead to a small increase of the ability/abilities. 
Besides, in the distribution $Beta(1,b)$, the larger $b$ is, the more the curve approaches $0$, which results in a higher chance in generating a smaller $\Delta \bm{\theta}$. 
It implies that a higher ability the learner has on either dimension, the harder for him/her to further improve the corresponding ability. 
Thus, $g_1(\bm{\theta}, \bm{a})$ and $g_2(\Delta \theta_1, \bm{\theta}, \bm{a})$ have positive coefficients in front of $\theta_1$ and $\theta_2$, respectively. 
Meanwhile, we assume that a higher ability on one dimension helps to increase the other dimension's ability, which results in a negative coefficient ahead of $\theta_2$ in $g_1(\bm{\theta}, \bm{a})$ and a negative coefficient ahead of $\theta_1$ in $g_2(\Delta \theta_1,\bm{\theta}, \bm{a})$.
In particular, assume the third learning material contains contents related to both abilities, and especially helps learners with intermediate or high ability level of addition to improve further on subtraction. 
This assumption is included in calculating $g_2(\bm{\theta}, \bm{a})$ when $\bm{a} = 3$ in equation \eqref{eq10}.
In addition, if the learner makes a big progress in mastering the ability of addition, there is a higher chance for the one to improve more on learning subtraction. 
Thus, the coefficient of $\Delta \theta_1$ in $g_2(\Delta \theta_1, \bm{\theta}, \bm{a})$ is negative which gives a curve that is less zero-concentrated as $\Delta \theta_1$ increases.
Consequently, $\Delta \theta_2$ has a higher possibility in increasing more as $\Delta \theta_1$ is large.
Note that the transition model is not required for adaptive learning system. 
The simulation gives an example in validating the model-free deep Q-learning algorithm in discovering the optimal learning policy. 

Estimation errors ranging from $1\%$ to $15\%$ are also added to estimated latent traits to evaluate their impacts on the adaptive learning system. 
Denote the estimation error vector by $\bm{e} = [e_1, e_2]^\top$, where $e_1$ and $e_2$ are generated by the same normal distribution such that $e_1, e_2 \sim \mathcal{N}(0,\sigma^2)$. 
As a result, $99.7\%$ of $e_1, e_2$ lie in the range of $(-3\sigma, 3\sigma)$. 
In the simulation, the estimated latent traits are calculated by the sum of the true latent traits and the estimation errors, which are $[\theta_1 + e_1, \theta_2 + e_2]^\top$. 
For instance, if the standard deviation $\sigma$ is $0.03$, the observation is $[\theta_1  + e_1, \theta_2 + e_2]^\top$, where $e_1, e_2 \sim \mathcal{N}(0, 0.03^2)$, and $99.7\%$ of $e_1, e_2$ lie in the range of $(-0.09, 0.09)$.

Two simulations cases are studied. 
In the first case, the DQN is trained against actual learners whose abilities' changes follow the MDP with kernel distributions described above. 
In this case, it is presumed that the optimal learning policy can be trained on sufficient number of learners. 
The resulting optimal learning policy is compared with a heuristic learning policy, which selects the next learning material that can improve the not-fully-mastered ability, and a random learning policy which selects any material randomly from the set of three. 
The impact of different estimation errors is also assessed. 
In the second case, the DQN is trained against an estimated transition model learning that is obtained using a small group of learners.
The resulting optimal learning policy is compared with that obtained by training against actual learners.

\subsection{Simulation Study I}

\begin{figure}[H]
\centering
\includegraphics[scale=0.6]{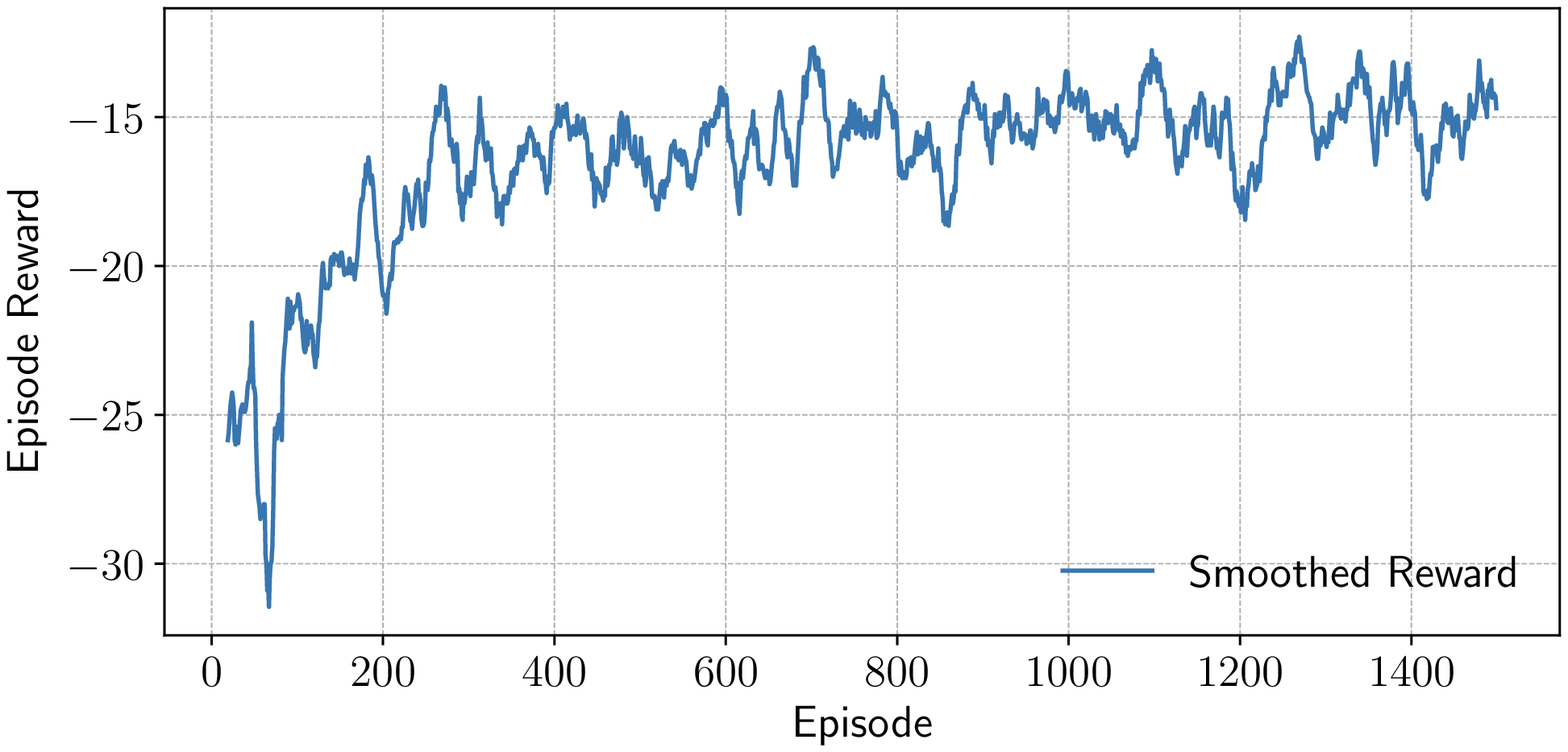}
\caption{Smoothed rewards under the deep Q-learning algorithm.\label{fig:rl_rewards}}
\end{figure}

Assume all learners are beginners on the two latent traits when using the adaptive learning system, i.e. $\bm{\Theta}^{(0)}=[0,0]^\top$.
The DQN has two hidden layers, the first of which has $64$ units and the second of which has $32$ units.
\blue{The DQN is trained against $2000$ learners that are simulated according to the method discussed earlier, i.e. $E = 2000$.}
Other parameters are chosen as follows: $\gamma = 0.9$, $\alpha = 6\times 10^{-4}$, $\overline{\epsilon} = 1.0$, $\underline{\epsilon} = 0.1$, $\tau_\epsilon = 2000$, $M = 256$.
The Adam algorithm is adopted for the training of the DQN.

Figure \ref{fig:rl_rewards} presents the smoothed reward under the deep Q-learning algorithm across the first $1500$ episodes with a smoothing window of $20$. 
\blue{It can be seen that the reward converges to $-15$ after $600$ episodes, which indicates the optimal learning policy is found after the DQN is trained using $600$ learners.}

\begin{figure}[H]
\centering
\includegraphics[scale=0.6]{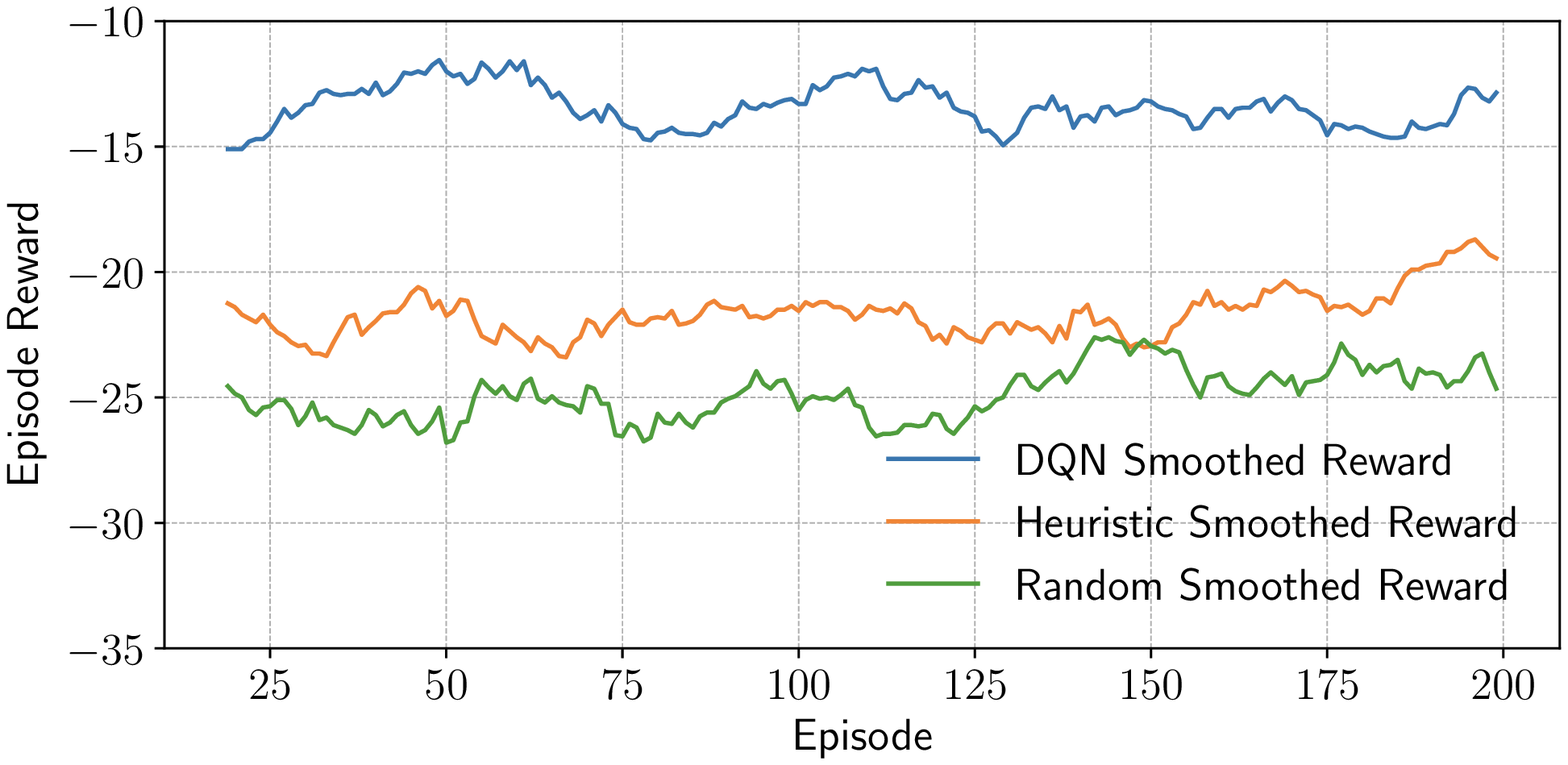}
\caption{Smoothed rewards under DQN, heuristic, and random learning policies.\label{fig:rl_rewards_compare}}
\end{figure}

\begin{table}
	\caption{Mean and Standard Deviation (SD) of Rewards under DQN, Heuristic, and Random Learning Policies. \label{T1}}
	\centering
	\begin{tabular}{cccc}
		\toprule
		Methods & DQN & Heuristic & Random\\
		\midrule
		Reward mean & -13.49 & -21.55 & -24.85\\
		Reward SD & 4.59 & 4.76 & 5.59\\
		\bottomrule
	\end{tabular}
\end{table}

Figure \ref{fig:rl_rewards_compare} and Table \ref{T1} compare smoothed rewards across \blue{$200$ new learners, labeled as episodes in Figure \ref{fig:rl_rewards_compare}}, with a smoothing window of $20$ between the optimal learning policy found by the deep Q-learning algorithm after being trained in $2000$ episodes---referred to as the DQN learning policy, the heuristic learning policy, and the random learning policy. 
\blue{The larger the reward is, the fewer learning steps a learner takes to fully master the two latent traits, or in another word, the better the learning policy is.}
Clearly, the rewards obtained by the deep Q-learning algorithm have a higher mean and smaller standard deviation (SD) than those obtained by the heuristic learning policy and the random learning policy. 
These results show that the learning policy found by the deep Q-learning algorithm is much better than the other two.

\begin{figure}[H]
\centering
\includegraphics[scale=0.6]{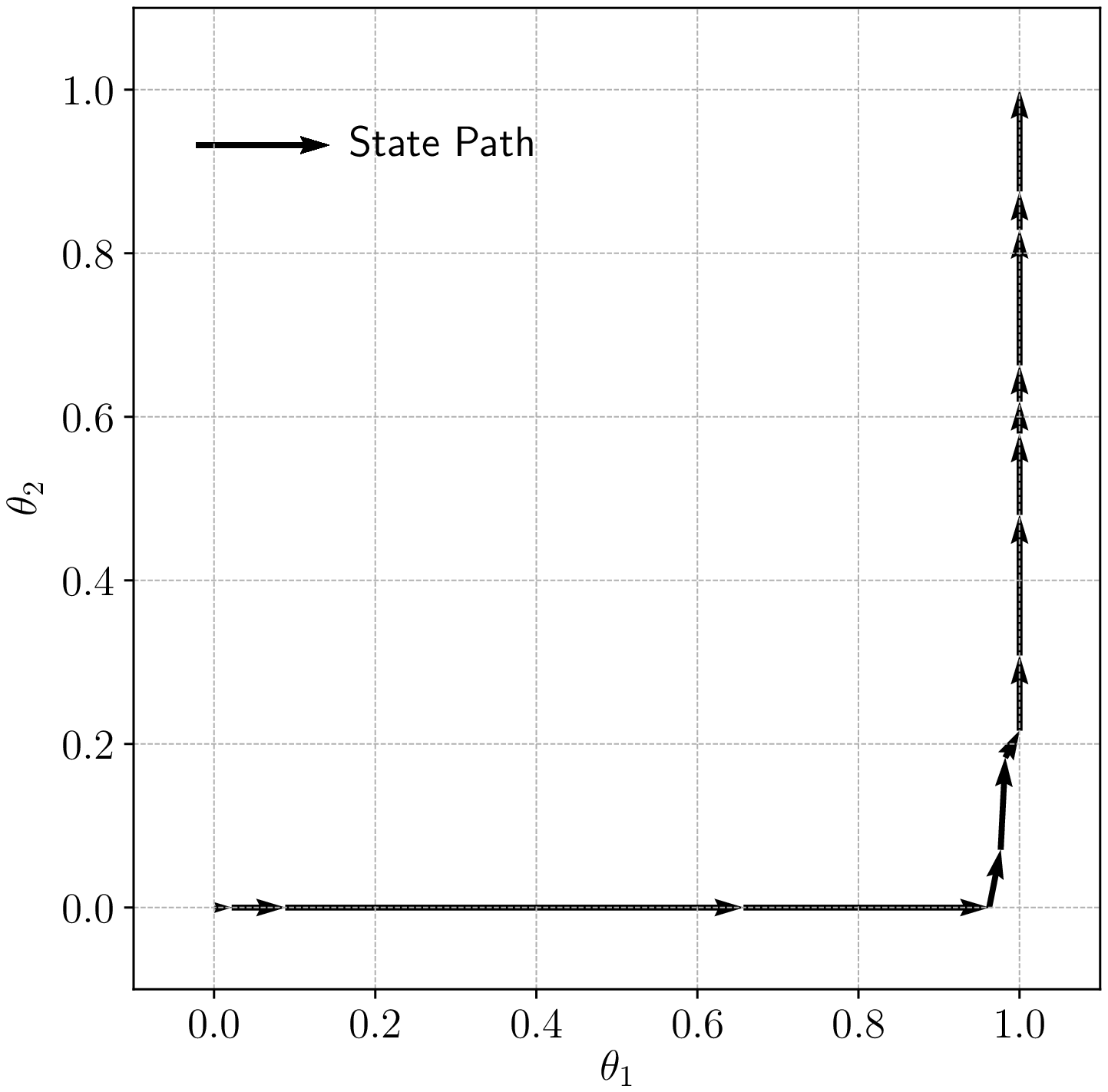}
\caption{An example of state transition path with action sequence of $1,1,1,1,3,3,3,3,3,2,2,2,2,2,2$.\label{fig:state_path}}
\end{figure}

Figure \ref{fig:state_path} presents an example of a state transition path that shows how the latent traits change with a sequence of actions taken under the DQN learning policy obtained without considering estimation error. 
Take the addition and subtraction test as an example. 
The first learning material is repeatedly selected to improve the learner's ability of addition at the beginning. 
Then the third material related to both addition and subtraction is selected. 
In the last few steps, the second learning material is chosen to further improve the learner's ability of subtraction. 

\begin{figure}[H]
\centering
\includegraphics[scale=0.6]{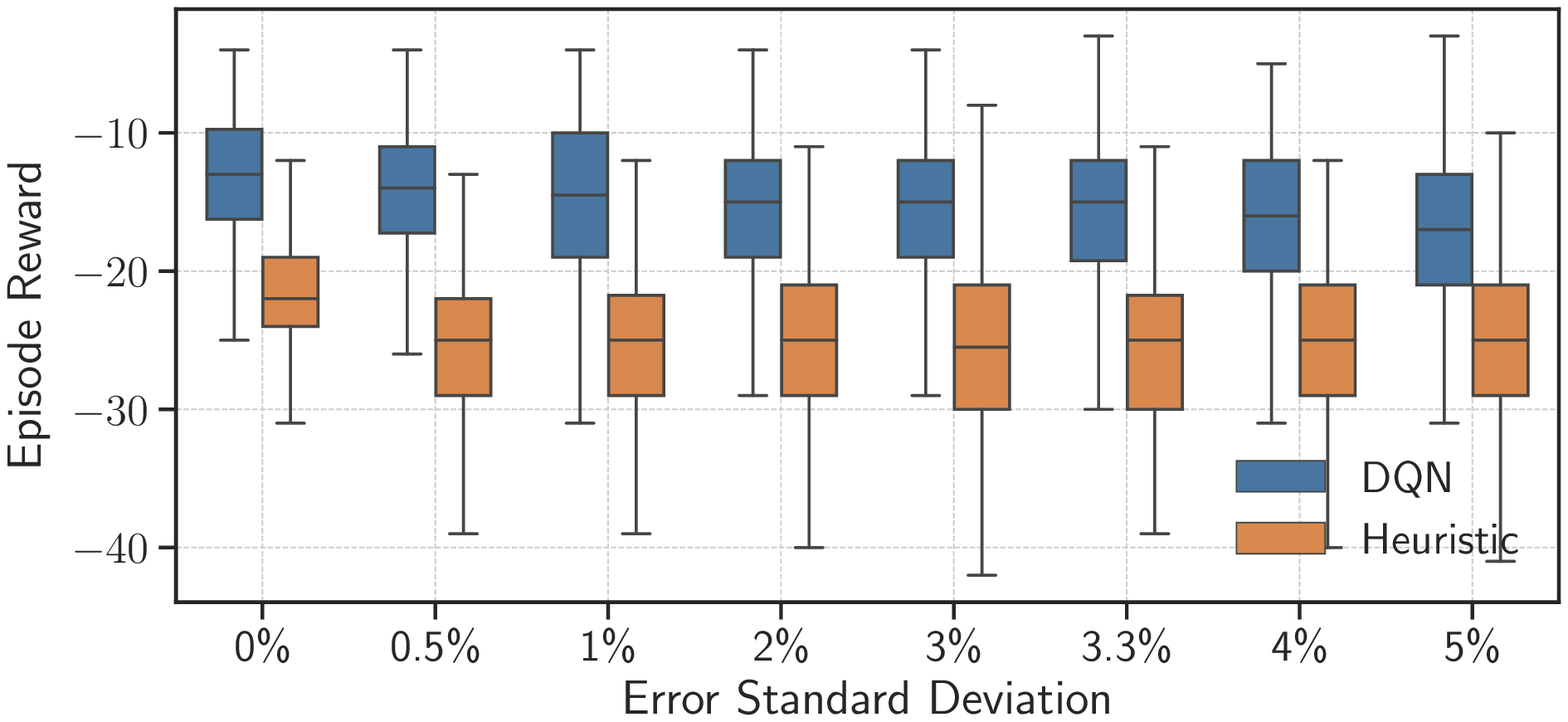}
\caption{Comparison of rewards under DQN and heuristic learning policies with various estimation errors.\label{fig:actual_error}}
\end{figure}

Figure \ref{fig:actual_error} compares rewards under the DQN and the heuristic learning policies when estimation errors with various standard deviations ($\sigma$) exist. 
It shows that the mean rewards obtained by the DQN learning policy under various estimation errors are consistently higher than those of the heuristic learning policy when estimation errors exist. 
That is, the DQN learning policy still outperforms the heuristic learning policy even with the presence of estimation errors, which demonstrates that the deep Q-learning algorithm is reliable and stable in finding optimal learning policy with the presence of estimation errors.

\subsection{Simulation Study II}

\blue{Next, we show the performance of the adaptive learning system with a transition model estimator, which is represented using a neural network with one hidden layer that has $32$ units.}
The prediction accuracy indices are presented in Table \ref{T2}. 
The train and test scores are defined as the coefficient of determination in the training and test sets respectively, calculated by
\begin{equation}\label{eq11}
	1 - \frac{\sum_{i=1}^H || \bm{s}^{(i)} - \hat{\bm{s}}^{(i)} ||^2}{\sum_{i=1}^H ||\bm{s}^{(i)} -\bm{\overline{s}} ||^2},
\end{equation}
where $\bm{s}$ is the true state, $\bm{\overline{s}}$ is average value of the true state, $\hat{\bm{s}}$ is the predicted state using previous state and the action taken, and $H$ is the number of the transitions. 
The best possible score is $1$.
The root mean square error (RMSE) is calculated by
\begin{equation}\label{eq12}
	RMSE = \sqrt{\dfrac{\sum_{i=1}^H || \bm{s}^{(i)} - \hat{\bm{s}}^{(i)} ||^2}{H}}.
\end{equation}

\begin{table}[!t]
	\caption{Accuracy of Transition Model Trained against Various Numbers of Learners. \label{T2}}
	\centering
	\begin{tabular}{cccccccccc}
		\toprule
		No. of learners & 10 & 20 & 30& 40& 50& 100& 150& 200& 2000\\
		\midrule
		Train Score & 0.96 & 0.97 & 0.97& 0.97& 0.97& 0.97& 0.97& 0.97& 0.97\\
		Test Score & 0.95 & 0.97& 0.96& 0.96& 0.97& 0.97& 0.97& 0.97& 0.97\\
		RMSE& 0.11 & 0.08& 0.09& 0.09& 0.08& 0.08& 0.08& 0.08& 0.08\\
		\bottomrule
	\end{tabular}
\end{table}

\begin{figure}[H]
\centering
\includegraphics[scale=0.6]{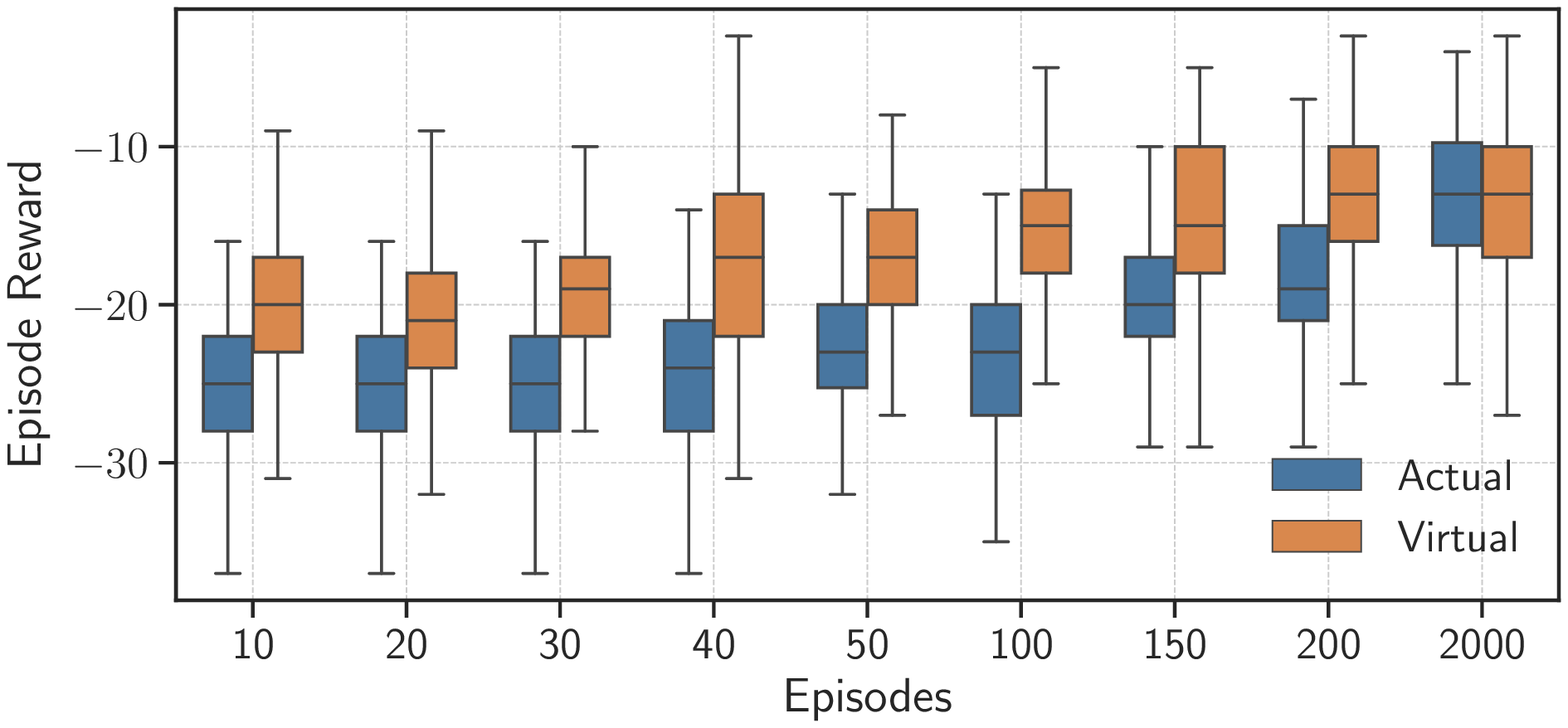}
\caption{Comparison of rewards under actual and virtual DQN learning policies.\label{fig:actual_virtual}}
\end{figure}

A DQN is trained on $2000$ episodes against the estimated transition model that is fitted using a certain number of actual learners; the learning policy corresponding to this DQN is referred to as the virtual DQN learning policy.
For the purpose of comparison, another DQN is trained on the same number of actual learners; the learning policy corresponding to this DQN is referred to as the actual DQN learning policy.
Essentially, these two learning policies differ in the way how the same set of actual learners are utilized.
\blue{The actual learners are simulated according to the method discussed in ``Simulation Overview" section and are used to train the actual DQN learning policy directly.
In contrast, these actual learners are used to first fit a transition model, which is then used to train the virtual DQN learning policy; this allows the virtual DQN learning policy to be trained over as many episodes as it needs.}
Figure \ref{fig:actual_virtual} compares rewards obtained by the two DQN learning policies when various numbers of actual learners are utilized.
It is shown that with no more than $200$ actual learners, the utilization of the transition model can significantly improve the performance of the learning policy, generating much larger mean rewards compared than the algorithm without using the transition model. 
When the number of learners is large enough, both two approaches found optimal learning policies and yield similar rewards.

%\begin{figure}[H]
%\centering
%\includegraphics[scale=0.6]{virtual_learn}
%\caption{\red{Smoothed rewards under the DQN algorithm over next 1950 episodes after being trained on 50 learners in the actual environment and after 50 learners fed into the neural network in the virtual environment.}\label{fig:virtual_learn}}
%\end{figure}
%
%Figure \ref{fig:virtual_learn} shows the comparison of the smoothed rewards with a smoothing window of $20$ over $1950$ episodes after $50$ learners are used to train the DQN and after $50$ learners are used to fit the transition model. 
%It is clear that the latter approach results in higher rewards at the beginning. 
%As the learning policy continues to be improved over more episodes, the mean rewards of these two approaches become closer. 
%It indicates that with the transition model, the adaptive learning system can generate a superior learning policy with only a small group of learners, and can also offer the optimal strategy as good as that being trained on the actual environment with more learners enrolled.

%%%%%%%%%%%%%%%%%%%%%%%%%%%%%%%%%%%%%%%%%%%%%%%%%%%%%%%%%%%%%%%%%%%%%%
\section{Concluding Remarks and Future Directions}

\blue{In this paper, we developed an MDP formulation for an adaptive learning system by describing learners' latent traits as continuous instead of simply classifying learners as mastery or non-mastery of certain skills.
The objective of the system is to improve learners' abilities to the prespecified target levels.}
We developed a deep Q-learning algorithm, which is a model-free DRL algorithm that can effectively find the optimal learning policy from data on learners' learning process without knowing the transition model of the learner's latent traits.
To cope with the challenge of insufficient state transition data, which may result in a poor performance of the deep Q-learning algorithm, we developed a transition model estimator that emulates the learner's learning process using neural networks, which can be used to further train the DQN and improve the its performance.

The two simulation studies presented in the paper verified that the proposed methodology is very efficient in finding a good learning policy for adaptive learning systems without any help from a teacher.
The optimal learning policy found by the DQN algorithm outperformed the heuristic and random methods with much higher rewards, or equivalently, much fewer learning steps/cycles for learners to reach the target levels of all prespecified abilities.
Particularly, with the aid of a transition model estimator, the adaptive learning system can find a good learning policy efficiently after training using a few learners.

The directions for extending the adaptive learning research include applying the adaptive learning system on actual learners to further assess the efficiency of the proposed methodology.
Both the DQN algorithm and the transition model estimator can be adopted and evaluated through real data analysis on an online learning platform.
Second, the adaptive learning system here consists of a latent trait estimator which uses measurement models to estimate latent traits and a learning policy.
Instead, some research construct the system assuming that learning materials influence learners' responses to test items directly, without the latent trait estimator incorporated \citep{lan2014time,lan2016contextual}.
As such, learners' learning process is modeled and traced directly and model-free algorithms can be proposed to find the optimal learning policy.
\blue{Third, because each group of learners assumes to follow a homogeneous MDP, further researches can be conducted to classify learners into groups before they use the adaptive learning system in order to find the optimal learning policy for each group.}
Fourth, machine learning algorithms for recommendation systems (e.g., collaborative filtering, matrix decomposition, etc.) can be incorporated with the DRL algorithm to better recommend not only optimal but also preferred materials to learners \citep{li2010contextual}. 
\blue{Fifth, it would be interesting to further examine how much better is the DQN learning policy than the random and heuristic learning policies under different scenarios and constrains.}
\blue{Sixth, it is also interesting to formulate the adaptive learning problem as a partially observable Markov decision process (POMDP) and explore solutions to the problem.}
Finally, future studies can include the modeled learning paths \citep{chen2018hidden,wang2018tracking} as learners' historical data to search the optimal learning policy more efficiently.

\vspace{\fill}\pagebreak

%% ITEM 9 [See the "howto.tex" file.]
%\appendix
%\renewcommand{\theequation}{A\arabic{equation}}
%\setcounter{equation}{0}
%\renewcommand{\thesection}{\Alph{subsection}}
%\setcounter{section}{0}
%\section*{Appendix}
%\section*{Appendix A}
%\section*{Appendix B}
%\vspace{\fill}\pagebreak

%% ITEM 10 [See the "howto.tex" file.]
%\begin{thebibliography}
%
%\bibitem
%
%\end{thebibliography}
\bibliographystyle{apalike}
\bibliography{ref.bib}

\begin{thebibliography}{}

\bibitem[Ackerman et~al., 2003]{ackerman2003using}
Ackerman, T.~A., Gierl, M.~J., and Walker, C.~M. (2003).
\newblock Using multidimensional item response theory to evaluate educational
  and psychological tests.
\newblock {\em Educational Measurement: Issues and Practice}, 22(3):37--51.

\bibitem[Bertsekas and Tsitsiklis, 1996]{bertsekas1996neuro}
Bertsekas, D.~P. and Tsitsiklis, J.~N. (1996).
\newblock {\em Neuro-dynamic programming}, volume~5.
\newblock Athena Scientific Belmont, MA.

\bibitem[Birnbaum, 1968]{birnbaum1968some}
Birnbaum, A. (1968).
\newblock Some latent trait models and their use in inferring an examinee's
  ability.
\newblock {\em Statistical theories of mental test scores}, pages 397--472.

\bibitem[Caicedo and Lazebnik, 2015]{caicedo2015active}
Caicedo, J.~C. and Lazebnik, S. (2015).
\newblock Active object localization with deep reinforcement learning.
\newblock In {\em Proceedings of the IEEE International Conference on Computer
  Vision}, pages 2488--2496.

\bibitem[Chang, 2015]{chang2015psychometrics}
Chang, H.-H. (2015).
\newblock Psychometrics behind computerized adaptive testing.
\newblock {\em Psychometrika}, 80(1):1--20.

\bibitem[Chen et~al., 2018a]{chen2018hidden}
Chen, Y., Culpepper, S.~A., Wang, S., and Douglas, J. (2018a).
\newblock A hidden markov model for learning trajectories in cognitive
  diagnosis with application to spatial rotation skills.
\newblock {\em Applied Psychological Measurement}, 42(1):5--23.

\bibitem[Chen et~al., 2018b]{chen2018recommendation}
Chen, Y., Li, X., Liu, J., and Ying, Z. (2018b).
\newblock Recommendation system for adaptive learning.
\newblock {\em Applied Psychological Measurement}, 42(1):24--41.

\bibitem[Firth, 1993]{firth1993bias}
Firth, D. (1993).
\newblock Bias reduction of maximum likelihood estimates.
\newblock {\em Biometrika}, 80(1):27--38.

\bibitem[Fran{\c{c}}ois-Lavet et~al., 2018]{franccois2018introduction}
Fran{\c{c}}ois-Lavet, V., Henderson, P., Islam, R., Bellemare, M.~G., Pineau,
  J., et~al. (2018).
\newblock An introduction to deep reinforcement learning.
\newblock {\em Foundations and Trends{\textregistered} in Machine Learning},
  11(3-4):219--354.

\bibitem[Goodfellow et~al., 2016]{goodfellow2016deep}
Goodfellow, I., Bengio, Y., and Courville, A. (2016).
\newblock {\em Deep learning}.
\newblock MIT press.

\bibitem[Gu et~al., 2017]{gu2017deep}
Gu, S., Holly, E., Lillicrap, T., and Levine, S. (2017).
\newblock Deep reinforcement learning for robotic manipulation with
  asynchronous off-policy updates.
\newblock In {\em 2017 IEEE International Conference on Robotics and Automation
  (ICRA)}, pages 3389--3396. IEEE.

\bibitem[Lan and Baraniuk, 2016]{lan2016contextual}
Lan, A.~S. and Baraniuk, R.~G. (2016).
\newblock A contextual bandits framework for personalized learning action
  selection.
\newblock In {\em Proceedings of the 9th International Conference on
  Educational Data Mining, Raleigh, NC: EDM}, pages 424--429.

\bibitem[Lan et~al., 2014]{lan2014time}
Lan, A.~S., Studer, C., and Baraniuk, R.~G. (2014).
\newblock Time-varying learning and content analytics via sparse factor
  analysis.
\newblock In {\em Proceedings of the 20th ACM SIGKDD international conference
  on Knowledge discovery and data mining}, pages 452--461. ACM.

\bibitem[Li et~al., 2010]{li2010contextual}
Li, L., Chu, W., Langford, J., and Schapire, R.~E. (2010).
\newblock A contextual-bandit approach to personalized news article
  recommendation.
\newblock In {\em Proceedings of the 19th international conference on World
  wide web}, pages 661--670. ACM.

\bibitem[Li et~al., 2018]{li2018optimal}
Li, X., Xu, H., Zhang, J., and Chang, H.-h. (2018).
\newblock Optimal hierarchical learning path design with reinforcement
  learning.
\newblock {\em arXiv preprint arXiv:1810.05347}.

\bibitem[Lord, 1980]{lord1980application}
Lord, F.~M. (1980).
\newblock Application of item response theory to practical testing problems.

\bibitem[Lord et~al., 1968]{lord1968statistical}
Lord, F.~M., Novick, M.~R., and Birnbaum, A. (1968).
\newblock Statistical theories of mental test scores. 1968.
\newblock {\em Reading: Addison-Wesley Google Scholar}.

\bibitem[Makransky and Glas, 2014]{makransky2014automatic}
Makransky, G. and Glas, C.~A. (2014).
\newblock An automatic online calibration design in adaptive testing.
\newblock {\em Journal of Applied Testing Technology}, 11(1):1--20.

\bibitem[Masters, 1982]{masters1982rasch}
Masters, G.~N. (1982).
\newblock A rasch model for partial credit scoring.
\newblock {\em Psychometrika}, 47(2):149--174.

\bibitem[McGlohen and Chang, 2008]{mcglohen2008combining}
McGlohen, M. and Chang, H.-H. (2008).
\newblock Combining computer adaptive testing technology with cognitively
  diagnostic assessment.
\newblock {\em Behavior Research Methods}, 40(3):808--821.

\bibitem[Mnih et~al., 2013]{mnih2013playing}
Mnih, V., Kavukcuoglu, K., Silver, D., Graves, A., Antonoglou, I., Wierstra,
  D., and Riedmiller, M. (2013).
\newblock {Playing atari with deep reinforcement learning}.
\newblock {\em arXiv preprint arXiv:1312.5602}.

\bibitem[Mnih et~al., 2015]{mnih2015human}
Mnih, V., Kavukcuoglu, K., Silver, D., Rusu, A.~A., Veness, J., Bellemare,
  M.~G., Graves, A., Riedmiller, M., Fidjeland, A.~K., Ostrovski, G., et~al.
  (2015).
\newblock {Human-level control through deep reinforcement learning}.
\newblock {\em Nature}, 518(7540):529.

\bibitem[Mulaik, 1972]{mulaik1972mathematical}
Mulaik, S. (1972).
\newblock A mathematical investigation of some multidimensional rasch models
  for psychological tests.
\newblock In {\em Annual Meeting of the Psychometric Society, Princeton, NJ}.

\bibitem[Muraki, 1992]{muraki1992generalized}
Muraki, E. (1992).
\newblock A generalized partial credit model: Application of an em algorithm.
\newblock {\em ETS Research Report Series}, 1992(1):i--30.

\bibitem[Rasch, 1960]{rash1960probabilistic}
Rasch, G. (1960).
\newblock Probabilistic models for some intelligence and attainment tests.
\newblock {\em Copenhagen: Danish Institute for Educational Research}.

\bibitem[Reckase, 1972]{reckase1972development}
Reckase, M.~D. (1972).
\newblock Development and application of a multivariate logistic latent trait
  model.

\bibitem[Reddy et~al., 2017]{reddy2017accelerating}
Reddy, S., Levine, S., and Dragan, A. (2017).
\newblock Accelerating human learning with deep reinforcement learning.
\newblock In {\em NIPS'17 Workshop: Teaching Machines, Robots, and Humans}.

\bibitem[Samejima, 1969]{samejima1969estimation}
Samejima, F. (1969).
\newblock Estimation of latent ability using a response pattern of graded
  scores.
\newblock {\em Psychometrika monograph supplement}.

\bibitem[Sutton and Barto, 2018]{sutton2018reinforcement}
Sutton, R.~S. and Barto, A.~G. (2018).
\newblock {\em Reinforcement learning: An introduction}.
\newblock MIT press.

\bibitem[Sympson, 1978]{sympson1978model}
Sympson, J.~B. (1978).
\newblock A model for testing with multidimensional items.
\newblock In {\em Proceedings of the 1977 computerized adaptive testing
  conference}, number 00014. University of Minnesota, Department of Psychology,
  Psychometric Methods.

\bibitem[Tang et~al., 2019]{tang2019reinforcement}
Tang, X., Chen, Y., Li, X., Liu, J., and Ying, Z. (2019).
\newblock A reinforcement learning approach to personalized learning
  recommendation systems.
\newblock {\em British Journal of Mathematical and Statistical Psychology},
  72(1):108--135.

\bibitem[Tseng and Hsu, 2001]{tseng2001multidimensional}
Tseng, F.-L. and Hsu, T.-C. (2001).
\newblock Multidimensional adaptive testing using the weighted likelihood
  estimation: a comparison of estimation methods.
\newblock In {\em Annual meeting of NCME, Seattle}.

\bibitem[Wang, 2015]{wang2015latent}
Wang, C. (2015).
\newblock On latent trait estimation in multidimensional compensatory item
  response models.
\newblock {\em Psychometrika}, 80(2):428--449.

\bibitem[Wang et~al., 2018]{wang2018tracking}
Wang, S., Yang, Y., Culpepper, S.~A., and Douglas, J.~A. (2018).
\newblock Tracking skill acquisition with cognitive diagnosis models: A
  higher-order, hidden markov model with covariates.
\newblock {\em Journal of Educational and Behavioral Statistics}, 43(1):57--87.

\bibitem[Warm, 1989]{warm1989weighted}
Warm, T.~A. (1989).
\newblock Weighted likelihood estimation of ability in item response theory.
\newblock {\em Psychometrika}, 54(3):427--450.

\bibitem[Weiss, 1982]{weiss1982improving}
Weiss, D.~J. (1982).
\newblock Improving measurement quality and efficiency with adaptive testing.
\newblock {\em Applied Psychological Measurement}, 6(4):473--492.

\bibitem[Whitely, 1980]{whitely1980multicomponent}
Whitely, S.~E. (1980).
\newblock Multicomponent latent trait models for ability tests.
\newblock {\em Psychometrika}, 45(4):479--494.

\bibitem[{Xu} et~al., 2019]{xu2019deep}
{Xu}, H., {Sun}, H., {Nikovski}, D., {Kitamura}, S., {Mori}, K., and
  {Hashimoto}, H. (2019).
\newblock Deep reinforcement learning for joint bidding and pricing of load
  serving entity.
\newblock {\em IEEE Transactions on Smart Grid}, pages 1--1.

\bibitem[Xu et~al., 2016]{xu2016personalized}
Xu, J., Xing, T., and Van Der~Schaar, M. (2016).
\newblock Personalized course sequence recommendations.
\newblock {\em IEEE Transactions on Signal Processing}, 64(20):5340--5352.

\bibitem[Zhang, 2013]{zhang2013procedure}
Zhang, J. (2013).
\newblock A procedure for dimensionality analyses of response data from various
  test designs.
\newblock {\em Psychometrika}, 78(1):37--58.

\bibitem[Zhang et~al., 2011]{zhang2011investigating}
Zhang, J., Xie, M., Song, X., and Lu, T. (2011).
\newblock Investigating the impact of uncertainty about item parameters on
  ability estimation.
\newblock {\em Psychometrika}, 76(1):97--118.

\bibitem[Zhang and Chang, 2016]{zhang2016smart}
Zhang, S. and Chang, H.-H. (2016).
\newblock From smart testing to smart learning: how testing technology can
  assist the new generation of education.
\newblock {\em International Journal of Smart Technology and Learning},
  1(1):67--92.

\end{thebibliography}

%% ITEM 11 [See the "howto.tex" file.]
%%%% You can put your Figures and Tables here
%%%% after the Reference Section.
%%%% BE SURE TO MARK IN THE TEXT WHERE
%%%% YOU WANT EACH FIGURE AND TABLE TO BE PLACED.
%%%% If you prefer, you can integrate your figures and tables into the text of your paper,
%%%% PROVIDED you will provide camera-ready copies of each figure.
%\vspace{\fill}\pagebreak
%\linespacing{1}

%\section*{Figures}
%
%\begin{figure}[h]
%\centerline{\includegraphics{figure01.eps}}
%\caption{Your figure caption goes here.}
%\end{figure}
%\vskip6pt

%\vspace{\fill}\pagebreak

%\section*{Tables}

%\vspace{\fill}\pagebreak

\end{document}